\documentclass{article} 
\usepackage{nips15submit_e,times}
\usepackage{hyperref}
\usepackage{url}

\usepackage{algorithm}
\usepackage{algorithmic}
\usepackage{amsfonts}
\usepackage{amsmath}
\usepackage[titletoc]{appendix}
\usepackage{bm}
\usepackage{dsfont} 
\usepackage{multicol}
\usepackage{multirow}
\usepackage{booktabs}
\usepackage{arydshln}
\usepackage[update,prepend]{epstopdf} 
\usepackage{gensymb}
\usepackage{wrapfig}
\usepackage{capt-of}

\usepackage{graphicx} 
\usepackage{subfigure}

\title{String Gaussian Process Kernels}

\author{
Yves-Laurent Kom Samo\\
Department of Engineering Science\\
Oxford-Man Institute of Quantitative Finance\\
University of Oxford\\
\texttt{ylks@robots.ox.ac.uk} \\
\And
Stephen Roberts\\
Department of Engineering Science\\
Oxford-Man Institute of Quantitative Finance\\
University of Oxford\\
\texttt{sjrob@robots.ox.ac.uk} \\
}

\nipsfinalcopy 
\begin{document}

\maketitle

\begin{abstract}
We introduce a new class of nonstationary kernels, which we derive as covariance functions of a novel family of stochastic processes we refer to as \textit{string Gaussian processes} (string GPs). We construct string GPs to allow for multiple types of local patterns in the data, while ensuring a mild global regularity condition. In this paper, we illustrate the efficacy of the approach using synthetic data and demonstrate that the model outperforms competing approaches on well studied, real-life datasets that exhibit nonstationary features.
\end{abstract}
\section{Introduction}
Over the past two decades, the use of kernels\footnote{We use the word kernel to denote any symmetric positive semi-definite function.} has been at the heart of many endeavours in the statistics and machine learning communities to extract hidden structures from datasets. Kernels are often used as a flexible way of departing from linear hypotheses in learning machines, thereby allowing for more complex nonlinear patterns \cite{vapnik95, vapnik98}. They have indeed been successfully applied to problems of classification, clustering, density estimation and regression. The duality between kernels and covariance functions has made kernels a critical tool for both frequentist and Bayesian statisticians. In the Bayesian nonparametrics community, kernels are often used as a covariance function of a Gaussian process (GP), introduced as a prior over a latent function. The family of covariance functions postulated for the GP is typically chosen so as to express prior domain knowledge about the underlying function, such as periodicity, regularity and range. The parameters of the kernel are then learned from the data. 

\textbf{Related work}

When one is concerned with automatically uncovering structures from the data, a flexible family of kernels should be used that can account for intricate patterns. Expressive kernels have been proposed to this end. In \cite{sparsespectrum, wilson2013gaussian}  two families of kernels were proposed that are dense in the class of stationary kernels, which allows flexibly learning the shape of the kernel from the data. Their approaches are however limited in that the learned kernel can be regarded as a global signature for the training data, that accounts simultaneously for every pattern/structure evidenced in the training data. As a result, prediction using these approaches will result in global extrapolation of local patterns or structures; that is prediction with the learned kernel will aim at replicating a mixture of all local patterns evidenced in the training dataset without any concept of spatial relevance. This might not be appropriate if the data exhibit multiple types of local patterns.

Varying patterns are often handled using nonstationary kernels. In \cite{paciorek2004nonstationary}  a method is proposed for constructing nonstationary covariance functions from any stationary one that involves introducing $n$ input dependent $d \times d$ covariance matrices to be inferred from the data. Their method however scales poorly with data size and input dimension, and as noted by the authors `it shows little advantage over stationary GP on held out data'. Another popular approach for introducing nonstationary kernels consists of postulating that stationarity holds in a new space resulting from a nonlinear deformation $\textbf{d}$ of the original input space (\cite{sampson92, damian01, Schmidt03}): $\textbf{k}(x, x^\prime) = f \big(\| \textbf{d}(x) - \textbf{d}(x^\prime) \|),$ where $f$ is positive definite. Although this idea is appealing, it relies on appropriately setting the deformation beforehand. In \cite{Schmidt03}  the deformation $\textbf{d}$ was learned in a fully nonparametric fashion, using a GP prior with identity mean and translation invariant covariance function. However, this effectively leads to a stationary kernel (see appendix). Another attempt to learn nonstationarity from the data in a fully Bayesian nonparametric way has been proposed by \cite{ggpm}. The authors considered functional priors of the form $y(x)= f(x) \exp{g(x)}$ where $f$ is a stationary GP and $g$ is a scaling function. For a given non-constant function $g$ such a prior yields a nonstationary Gaussian process. However, when a stationary GP prior is put on the function $g$ as in \cite{ggpm} the resulting functional prior $y(x)= f(x) \exp{g(x)}$ becomes stationary, with a covariance function that can be determined analytically (see appendix).

Other authors have introduced nonstationarity by noting that in most cases, heterogeneous patterns in the data are locally homogeneous. For example, \cite{kim} partitioned the domain using Voronoi tessellations and considered independent stationary GPs defined on elements of the partition and \cite{gramacy} used tree based partitioning and postulated independent stationary GPs on the leaves. The two methods can also be regarded as mixtures of GP experts (\cite{tresp, Rasmussen01infinitemixtures, Meeds_analternative, icml2013_ross13a}) on non-overlapping domains. This approach has been shown to be very flexible, and is truly nonstationary. It also scales considerably better than alternative approaches such as \cite{paciorek2004nonstationary}, as it has complexity $\mathcal{O}(\sum_{k} n_k^3)$ ---which is better than the typical GP complexity $\mathcal{O}\big((\sum_{k} n_k)^3\big)$--- and it introduces fewer additional parameters. However, these methods present two major pitfalls.  Firstly, the independence between experts results in higher posterior uncertainty, as individual experts have fewer data to learn from. More importantly, the resulting functional priors are discontinuous at the boundary of the partition, which might not be appropriate if the latent function to be inferred is known to be smooth.

The approach we present in this paper improves on the above work in that we construct a new stochastic process resulting from a collaboration between local GP experts on non-overlapping domains. Each expert is driven by its own arbitrarily flexible kernel, and shares boundary conditions with neighbours, conditional on which they are independent. This offers principled nonstationary construction from existing kernels, while ensuring the functional prior is continuously differentiable. The rest of the paper is structured as follows. In section \ref{sct:sk} we construct \textit{string Gaussian processes} indexed on $\mathbb{R}$, we derive their covariance functions and provide multivariate extensions. In section \ref{sct:exps}, we illustrate the necessity for our approach on synthetic data and demonstrate that it outperforms alternative models on well studied real-life datasets. We conclude with a discussion in section \ref{sct:dsc}.
\section{String GP Kernels}
\label{sct:sk}
In this section we construct a new class of stochastic processes whose covariance functions will give rise to \textit{string GP kernels}. First we consider the joint between a GP and its derivative (when it exists).
\vspace{-1em}
\begin{proposition}
\label{prop:derivative_processes}
Let $I$ be an interval, $\textbf{k}: I \times I \rightarrow \mathbb{R}$ a $\mathcal{C}^2$ symmetric positive definite function\footnote{$\mathcal{C}^1$ (resp. $\mathcal{C}^2$) functions denote functions that are once (resp. twice) continuously differentiable on their domains.}, $m: I \rightarrow \mathbb{R}$ a $\mathcal{C}^1$ function. There exists a $\mathbb{R}^2$ valued stochastic process $\big(D_t\big)_{t \in I}, D_t=(z_t, z_t^\prime)$, such that for all $t_1, \dots, t_n \in I$, $(z_{t_1}, \dots, z_{t_n}, z_{t_1}^\prime, \dots, z_{t_n}^\prime)$ is a Gaussian vector with mean $(m(t_1), \dots, m(t_n), $$\frac{\text{d}m}{\text{dt}}(t_1), \dots, \frac{\text{d}m}{\text{dt}}(t_n))$ and covariance matrix such that $~~\text{cov}(z_{t_i}, z_{t_j})=\textbf{k}(t_i, t_j),~~$ $\text{cov}(z_{t_i}, z_{t_j}^\prime)=\frac{\partial \textbf{k}}{\partial y }(t_i, t_j)$ and $\text{cov}(z_{t_i}^\prime, z_{t_j}^\prime)=\frac{\partial^2 \textbf{k}}{\partial x \partial y }(t_i, t_j).$ We herein refer to $(D_t)_{t \in I}$ as a \textbf{derivative Gaussian process} (DGP). Moreover, $(z_t)_{t \in I}$ is a GP with mean function $m$, covariance function $\textbf{k}$ and that is $\mathcal{C}^1$ in the $L^2$ (mean square) sense. Furthermore, $(z^\prime_t)_{t \in I}$ is a GP with mean function $\frac{\text{d}m}{\text{dt}}$ and covariance function $\frac{\partial^2 \textbf{k}}{\partial x \partial y }$, and $(z^\prime_t)_{t \in I}$ is the $L^2$ derivative of the process $(z_t)_{t \in I}$. \textbf{Proof}: see appendix.
\end{proposition}
We consider a kernel $\textbf{k}$ as \textbf{degenerate at} $a$ when a DGP $(z_t, z_t^\prime)_{t \in I}$ with kernel $\textbf{k}$ is such that $z_a$ and $z_a^\prime$ are perfectly correlated\footnote{Or equivalently when the Gaussian vector $(z_a, z_a^\prime)$ is degenerate.}: $\vert \text{corr}(z_a, z_a^\prime) \vert= 1$. As an example, the linear kernel $\textbf{k}(u,v) = \sigma^2(u-c)(v-c)$ is degenerate at $0$. Moreover, we will consider a kernel $\textbf{k}$ as \textbf{degenerate at} $b$ \textbf{given} $a$ when it is not degenerate at $a$ and when a DGP $(z_t, z_t^\prime)_{t \in I}$ with kernel $\textbf{k}$ is such that the variances of $z_b$ and $z_b^\prime$ conditional on $(z_a, z_a^\prime)$ are both zero\footnote{That is when the Gaussian vector $(z_a, z_a^\prime)$ is not degenerate but $(z_a, z_a^\prime, z_b, z_b^\prime)$ is.}. For instance, the periodic kernel proposed by \cite{gp_intro} with period $T$ is degenerate at $u+T$ given $u$. 

An important subclass of DGPs in our construction are the processes resulting from conditioning paths of a DGP to take specific values at certain times $(t_1, \dots, t_c)$. We herein refer to those processes as \textbf{\textit{conditional derivative Gaussian processes}} (CDGP). As an illustration, when $\textbf{k}$ is $\mathcal{C}^3$ on $I \times I$ with $I=[a,b]$, and neither degenerate at $a$ nor degenerate at $b$ given $a$,  the CDGP on $I=[a,b]$ with unconditional mean function $m$ and unconditional covariance function $\textbf{k}$ that  is conditioned to start at $(\tilde{z}_a, \tilde{z}_a^\prime)$ and to end at $(\tilde{z}_b, \tilde{z}_b^\prime)$, is the DGP whose mean and covariance functions read:
\begin{align}
&m_c^{a, b}(t) = m(t) +  \tilde{\textbf{K}}_{t; (a, b)} \textbf{K}_{(a,b); (a, b)}^{-1}  \begin{bmatrix} \tilde{z}_a - m(a) & \tilde{z}_a^\prime - \frac{d m}{dt}(a) & \tilde{z}_b - m(b) & \tilde{z}_b^\prime - \frac{d m}{dt}(b)\end{bmatrix}^T, \nonumber
\end{align}
\begin{align}
\label{eq:double_cond_cov}
\textbf{k}_c^{a,b}(t, s) =  \textbf{k}(t, s) - \tilde{\textbf{K}}_{t; (a, b)} \textbf{K}_{(a,b); (a, b)}^{-1}  \tilde{\textbf{K}}_{s; (a, b)}^T,\nonumber
\end{align}
\[\text{where}
~~\tilde{\textbf{K}}_{t; a} =
\begin{bmatrix} 
\textbf{k}(t, a) & \frac{\partial \textbf{k}}{\partial y}(t, a) 
\end{bmatrix},
~~\tilde{\textbf{K}}_{t; (a, b)} = 
\begin{bmatrix} 
\tilde{\textbf{K}}_{t; a} & \tilde{\textbf{K}}_{t; b}
\end{bmatrix},
~~\textbf{K}_{u; v} = 
\begin{bmatrix} 
\textbf{k}(u, v) & \frac{\partial \textbf{k}}{\partial y}(u, v) \\
\frac{\partial \textbf{k}}{\partial x}(u, v)  & \frac{\partial^2 \textbf{k}}{\partial x \partial y}(u, v) 
\end{bmatrix},\]
and $
\textbf{K}_{(a, b); (a, b)} = 
\begin{bmatrix} 
\textbf{K}_{a; a} & \textbf{K}_{a; b} \\
\textbf{K}_{b; a} & \textbf{K}_{b; b}
\end{bmatrix}$. It is worth noting that both $\textbf{K}_{a; a}$ and $\textbf{K}_{(a,b); (a, b)}$ are indeed invertible because the kernel is assumed to be neither degenerate at $a$ nor degenerate at $b$ given $a$. Hence, the support of $(z_a, z_a^\prime, z_b, z_b^\prime)$ is $\mathbb{R}^4$, and any values can be used for conditioning.

\textbf{\underline{String Gaussian processes}}

\begin{figure}[ht]
\begin{center}
\vspace{-1em}
\centerline{(a)\includegraphics[width=0.47\textwidth]{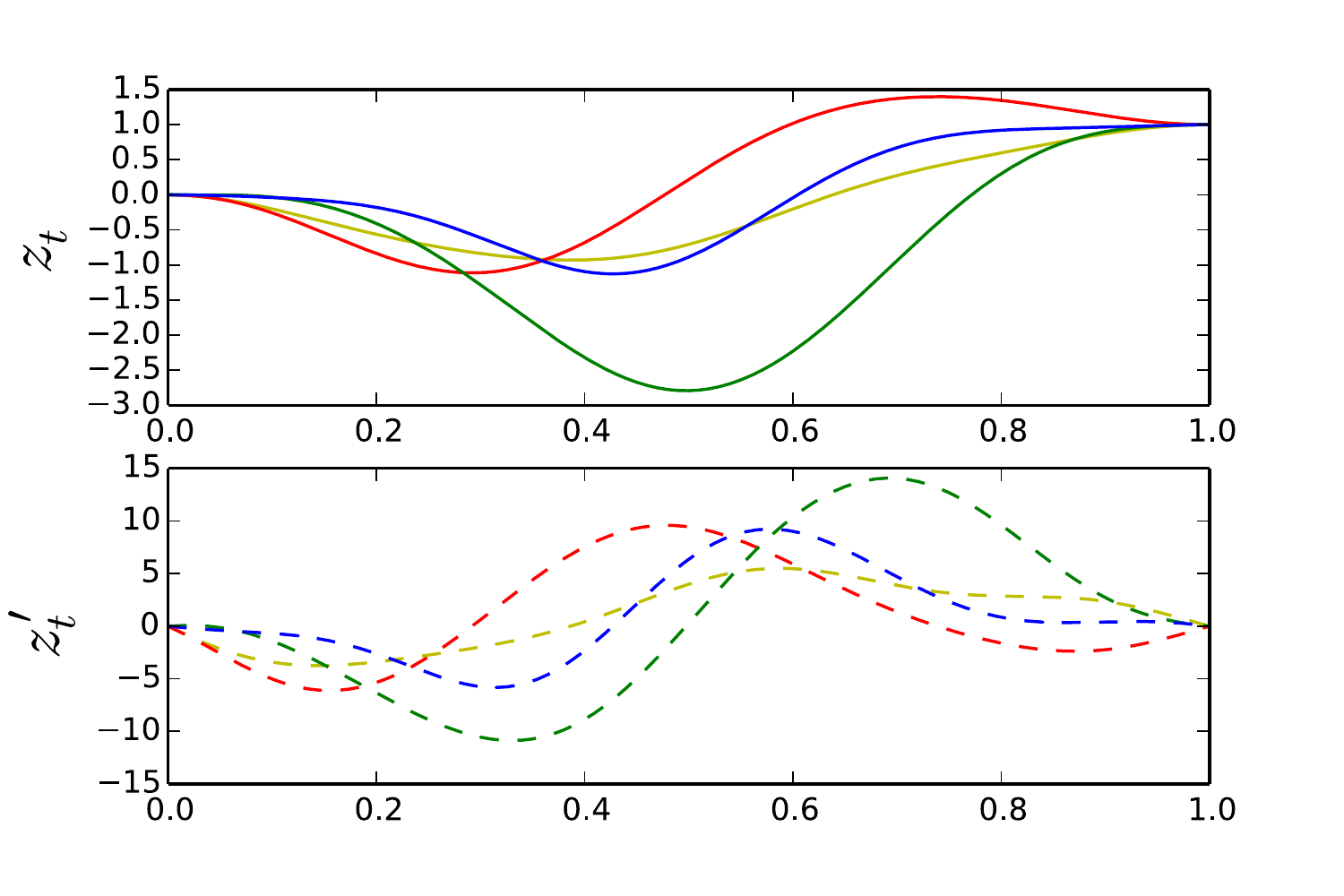}(b)\includegraphics[width=0.47\textwidth]{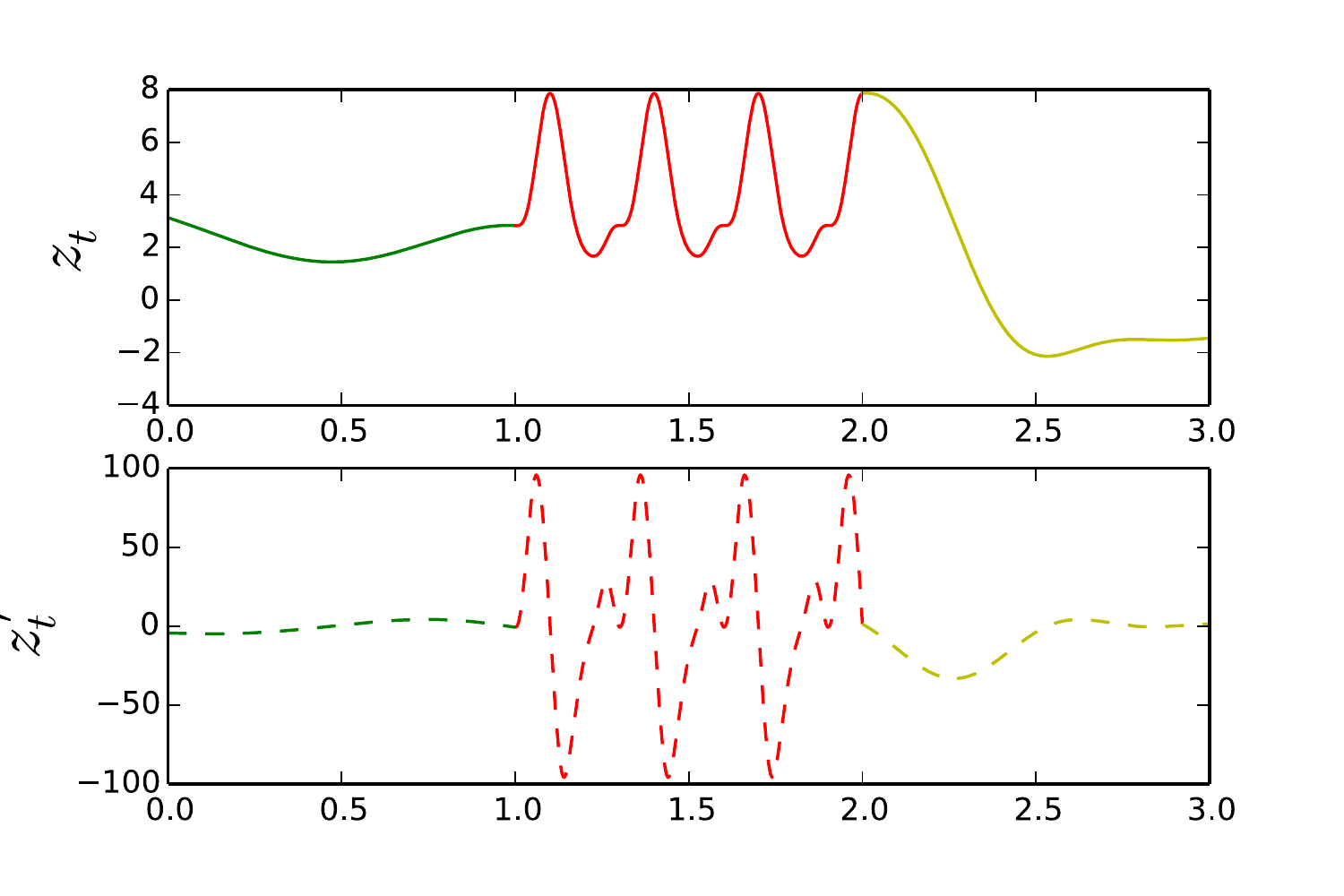}}
\vspace{-1em}
\caption{Examples of CDGP (a) and string DGP (b).}
\vspace{-1.5em}
\label{fig:example_plots}
\end{center}
\end{figure}

The following theorem, at the core of our framework, establishes that it is possible to connect together GPs on a partition of an interval $I$, in a manner consistent enough that the newly constructed stochastic object will be a stochastic process on $I$ and in a manner restrictive enough that any two connected GPs will share just enough information to ensure that the constructed stochastic process is continuously differentiable ($\mathcal{C}^1$) on $I$ in the $L^2$ sense.
\begin{theorem}\label{theo:sgp} Let $a_0<\dots<a_K$, $I=[a_0, a_K]$ and let $p_\mathcal{N}(x; \mu, \Sigma)$ be the multivariate Gaussian density with mean vector $\mu$ and covariance matrix $\Sigma$. Furthermore, let $(m_k:  [a_{k-1}, a_k] \to \mathbb{R})_{k \in [1..K]}$ be $\mathcal{C}^1$ functions, and   $(\textbf{k}_k: [a_{k-1}, a_k] \times  [a_{k-1}, a_k]\to \mathbb{R})_{k \in [1..K]}$ be $\mathcal{C}^3$ symmetric positive definite functions, neither degenerate at $a_{k-1}$, nor degenerate at $a_k$ given $a_{k-1}$.

(A) There exists an $\mathbb{R}^2$ valued stochastic process $(SD_{t})_{t \in I}, ~ SD_t=(z_t, z_t^\prime)$  such that:\\
1) The probability density of $(SD_{a_0}, \dots, SD_{a_K})$ reads: 
$p_{b}(x_0, \dots, x_K) := \prod_{k=0}^K p_\mathcal{N}\big(x_k;  \mu^b_k, \Sigma_k^b\big)$.
\begin{equation}
\label{eq:sb0k}
\text{where:}~~~~
\Sigma_0^b = {}_1\textbf{K}_{a_0; a_0}, ~~
\forall ~ k>0 ~~\Sigma_k^b = {}_k\textbf{K}_{a_k; a_k} - {}_k\textbf{K}_{a_k; a_{k-1}}~{}_k\textbf{K}_{a_{k-1}; a_{k-1}}^{-1}~{}_k\textbf{K}_{a_k; a_{k-1}}^T,
\end{equation}
\begin{equation}
\label{eq:mb0k}
\mu_0^b={}_1\textbf{M}_{a_0}, ~~
\forall ~ k>0 ~~\mu^b_k={}_k\textbf{M}_{a_k} + {}_k\textbf{K}_{a_k; a_{k-1}}~{}_k\textbf{K}_{a_{k-1}; a_{k-1}}^{-1}(x_{k-1}-{}_k\textbf{M}_{a_{k-1}}), 
\end{equation}
\[
\text{with}~~~~{}_k\textbf{K}_{u;v} = 
\begin{bmatrix} 
\textbf{k}_k(u, v) & \frac{\partial \textbf{k}_k}{\partial y}(u, v) \\
\frac{\partial \textbf{k}_k}{\partial x}(u, v)  & \frac{\partial^2 \textbf{k}_k}{\partial x \partial y}(u, v) 
\end{bmatrix}, ~~~~\text{and}~~~~
{}_k\textbf{M}_u =\begin{bmatrix} m_k(u) \\ \frac{d m_k}{dt}(u) \end{bmatrix}.
\]
2) Conditional on $(SD_{a_k} = x_k)_{k \in [0..K]}$, the restrictions $(SD_{t})_{t \in  ]a_{k-1}, a_k[},~ k \in [1..K]$ are \textbf{independent CDGPs}, respectively with unconditional mean function $m_k$ and unconditional covariance function $\textbf{k}_k$ and that are conditioned to take values $x_{k-1}$ and $x_k$ at $a_{k-1}$ and $a_k$ respectively.
We refer to $(SD_{t})_{t \in I}$  as a \textbf{string derivative Gaussian process} (string DGP), and to its first coordinate $(z_{t})_{t \in I}$ as a \textbf{string Gaussian process} (string GP) namely, \[(z_{t})_{t \in I} \sim \mathcal{SGP}(\{a_k\}, \{m_k\}, \{\textbf{k}_k\}).\]
(B) The \textbf{string Gaussian process} $(z_t)_{t \in I}$ defined in (A) is a Gaussian process that is $\mathcal{C}^1$ in the $L^2$ sense and its $L^2$ derivative is the process $(z_t^\prime)_{t \in I}$ defined in (A). \textbf{Proof}: see appendix.
\end{theorem}
In our \textit{collaborative local GP experts} analogy, Theorem \ref{theo:sgp} stipulates that each local expert (string) takes as message from the previous expert its left hand side boundary conditions, conditional on which it generates its right hand side boundary conditions, which it then passes on to the next expert. Conditional on their boundary conditions local experts are independent of each other, and resemble vibrating pieces of string on fixed extremities, hence the name \textit{string Gaussian process}. Figure \ref{fig:example_plots}(a) illustrates 3 independent draws from a CDGP conditioned to start at $0$ with derivative $0$ and to finish at $1$ with derivative $0$. The unconditional GP used was stationary with mean $0$ and squared exponential covariance function with variance $1$ and input scale $0.2$. Figure \ref{fig:example_plots}(b) illustrates a draw from a string GP $(z_t)$ with 3 strings and its derivative $(z_t^\prime)$, under squared exponential kernels (green and yellow strings), and the periodic kernel of \cite{gp_intro} (red string).

\textbf{\underline{Univariate String GP Kernels}}

We call \textit{\textbf{string GP kernel}} the covariance function of a \textit{string GP}. As we do not impose any restriction on the functional form of the unconditional kernels driving strings, it is not possible to derive a functional form for \textit{string GP kernels}, but they are readily available from unconditional kernels. We now derive how to evaluate the joint covariance structure of a string GP and its derivative, 
\[\bar{\textbf{K}}_{u; v} :=
\begin{bmatrix} 
\text{cov}(z_u, z_v) & \text{cov}(z_u, z_v^{\prime}) \\ 
\text{cov}(z_u^{\prime}, z_v)  & \text{cov}(z_u^{\prime}, z_v^{\prime})
\end{bmatrix}=
\begin{bmatrix} 
k_{SGP}(u,v) & \frac{\partial k_{SGP}}{\partial v}(u,v) \\ 
\frac{\partial k_{SGP}}{\partial u}(u,v)  & \frac{\partial^2 k_{SGP}}{\partial u \partial v}(u,v) \end{bmatrix}=
\bar{\textbf{K}}_{v; u}^T,
\] 
which corresponds to \textit{string GP kernels} and their partial derivatives. To do so, we need the following lemma:
\begin{lemma}
\label{lem:gaussian_message}
Let $X$ be a multivariate Gaussian with mean $\mu_X$ and covariance matrix $\Sigma_X$. If conditional on $X$, $Y$ is a multivariate Gaussian with mean $MX + A$  and covariance matrix $\Sigma_Y^c$ where $M$, $A$ and $\Sigma_Y^c$ do not depend on $X$, then $(X, Y)$ is a jointly Gaussian vector  \[\text{with mean }\mu_{X;Y}=\begin{bmatrix} \mu_X \\ M\mu_X + A \end{bmatrix} \text{ and covariance matrix } \Sigma_{X;Y}=\begin{bmatrix} \Sigma_X & \Sigma_X M^T \\ M\Sigma_X & \Sigma_Y^c + M \Sigma_X M^T\end{bmatrix}.\]
\textbf{Proof}: see appendix.
\end{lemma}

To derive $\bar{\textbf{K}}_{u; v}$, we start by noting that the covariance function of a string GP is the same as that of another string GP whose strings have the same unconditional kernels but unconditional mean functions $m_k=0$, so that to evaluate \textit{string GP kernels} we may assume that $\forall k, ~m_k=0$ without loss of generality. We deal with the case where $u$ and $v$ are both boundary times, after which we will generalise to other times.

\textit{\textbf{Boundary times}}: We note from Theorem \ref{theo:sgp} that the restriction $(z_t, z_t^{\prime})_{t \in  [a_0, a_{1}]}$ is the DGP with mean $0$ and covariance function $\textbf{k}_1$. Thus, $(z_{a_0}, z_{a_0}^{\prime}, z_{a_1}, z_{a_1}^{\prime})$ is jointly Gaussian, and if we let ${}_k\textbf{K}_{u; v}$ be as in Theorem \ref{theo:sgp},
\begin{align}
\boxed{\bar{\textbf{K}}_{a_0; a_0} = {}_1\textbf{K}_{a_0; a_0}, ~~ \bar{\textbf{K}}_{a_1; a_1} = {}_1\textbf{K}_{a_1; a_1}, ~~ \bar{\textbf{K}}_{a_0; a_1} = {}_1\textbf{K}_{a_0; a_1}.}
\end{align}
We recall that conditional on the boundary conditions at or prior to $a_{k-1}$, $(z_{a_k}, z_{a_k}^{\prime})$ is Gaussian with mean 
${}^b_kM
\begin{bmatrix}
z_{a_{k-1}} &
z^{\prime}_{a_{k-1}}
\end{bmatrix}^T~~\text{where}~~
{}^b_kM= {}_k\textbf{K}_{a_{k}; a_{k-1}}~_k\textbf{K}_{a_{k-1}; a_{k-1}}^{-1},
$ and covariance matrix $
{}^b_k\Sigma =  {}_k\textbf{K}_{a_k; a_k} - {}^b_kM ~_k\textbf{K}_{a_{k-1}; a_k}.
$ 
Hence using Lemma \ref{lem:gaussian_message} with $M=\begin{bmatrix}{}^b_kM & 0 & \dots & 0\end{bmatrix}$ where there are $(k-1)$ null block $2 \times 2$ matrices, and noting that $(z_{a_0}, z_{a_0}^{\prime}, \dots, z_{a_{k-1}}, z_{a_{k-1}}^{\prime})$ is jointly Gaussian, it follows that $(z_{a_0}, z_{a_0}^{\prime}, \dots, z_{a_k}, z_{a_k}^{\prime})$ is jointly Gaussian, that $(z_{a_k}, z_{a_k}^{\prime})$ has covariance matrix
\begin{align}
\boxed{\bar{\textbf{K}}_{a_k; a_k} = {}^b_k\Sigma + {}^b_kM~\bar{\textbf{K}}_{a_{k-1}; a_{k-1}}~{}^b_kM^T,}
\end{align}
and that the covariance matrix between the boundary conditions at $a_k$ and at any earlier boundary time $a_l, ~ l<k$ reads:
\begin{align}
 \boxed{\bar{\textbf{K}}_{a_k; a_l} = {}^b_kM~\bar{\textbf{K}}_{a_{k-1}; a_l}.}
\end{align}
\textbf{\textit{String times}}:  Let $u \in [a_{p-1}, a_p], ~ v\in [a_{q-1}, a_q]$. By the law of total expectation, we have that
\[
\bar{\textbf{K}}_{u; v} := \text{E}\bigg( 
\begin{bmatrix}
z_{u} \\
z^{\prime}_{u}
\end{bmatrix}
\begin{bmatrix}
z_{v} & z^{\prime}_{v}
\end{bmatrix}\bigg) =
\text{E}\Bigg( \text{E}\bigg( 
\begin{bmatrix}
z_{u} \\
z^{\prime}_{u}
\end{bmatrix}
\begin{bmatrix}
z_{v} & z^{\prime}_{v}
\end{bmatrix}\bigg| \mathcal{B}(p, q)\bigg) \Bigg),
\]
where $\mathcal{B}(p, q)$ refers to the boundary conditions at the boundaries of the $p$-th and $q$-th strings, in other words $\bigg\{z_{x}, z^{\prime}_{x}, ~x \in \big\{a_{p-1}, a_p,a_{q-1}, a_q \big\}\bigg\}$. Furthermore,  using the definition of  the covariance matrix under the conditional law, it follows that
\begin{equation}
\label{eq:tot_kern_cond_cov}
\text{E}\bigg( 
\begin{bmatrix}
z_{u} \\
z^{\prime}_{u}
\end{bmatrix}
\begin{bmatrix}
z_{v} & z^{\prime}_{v}
\end{bmatrix}\bigg| \mathcal{B}(p, q)\bigg) = {}_c\bar{\textbf{K}}_{u; v} + \text{E}\bigg( 
\begin{bmatrix}
z_{u} \\
z^{\prime}_{u}
\end{bmatrix}
\bigg| \mathcal{B}(p, q)\bigg)
 \text{E}\bigg( 
\begin{bmatrix}
z_{v} & z^{\prime}_{v}
\end{bmatrix}\bigg| \mathcal{B}(p, q)\bigg),
\end{equation}
where $_c\bar{\textbf{K}}_{u; v}$ refers to the covariance matrix between $(z_u, z^{\prime}_u)$ and $(z_v, z^{\prime}_v)$ conditional on the boundary conditions $\mathcal{B}(p, q)$ and can be easily evaluated from Theorem \ref{theo:sgp}. In particular,
\begin{equation}
\label{eq:kern_cond_cov}
\text{if}~ p \neq q, ~_c\bar{\textbf{K}}_{u; v} =0, ~\text{and if}~ p=q, ~_c\bar{\textbf{K}}_{u; v} = {}_p\textbf{K}_{u;v} - {}_p\Lambda_u
\begin{bmatrix}
_p\textbf{K}_{v;a_{p-1}}^T 
\\  _p\textbf{K}_{v;a_p}^T 
\end{bmatrix},
\end{equation}
where 
\[\forall ~ x, l, ~_l\Lambda_x = 
\begin{bmatrix}
_l\textbf{K}_{x;a_{l-1}} &  _l\textbf{K}_{x;a_l} 
\end{bmatrix} 
\begin{bmatrix} 
_l\textbf{K}_{a_{l-1};a_{l-1}} &  _l\textbf{K}_{a_{l-1};a_l} \\
_l\textbf{K}_{a_l;a_{l-1}} &  _l\textbf{K}_{a_l;a_l}
\end{bmatrix}^{-1}.\]
We also note that \[
\text{E}\bigg( 
\begin{bmatrix}
z_{u} \\
z^{\prime}_{u}
\end{bmatrix}
\bigg| \mathcal{B}(p, q)\bigg) = {}_p\Lambda_u 
\begin{bmatrix}
z_{a_{p-1}} \\
z^{\prime}_{a_{p-1}} \\
z_{a_p} \\
z^{\prime}_{a_p} \\
\end{bmatrix}~\text{and}~~
\text{E}\bigg( 
\begin{bmatrix}
z_{v} & z^{\prime}_{v}
\end{bmatrix}
\bigg| \mathcal{B}(p, q)\bigg) = 
\begin{bmatrix}
z_{a_{q-1}} & z^{\prime}_{a_{q-1}} & z_{a_q} & z^{\prime}_{a_q} 
\end{bmatrix}
{}_q\Lambda_v ^T.
\]
Hence, taking the expectation with respect to the boundary conditions on both sides of Equation (\ref{eq:tot_kern_cond_cov}), we obtain:
\begin{equation}
\boxed{
\forall ~u \in [a_{p-1}, a_{p}], ~  v \in [a_{q-1}, a_{q}],~~
\bar{\textbf{K}}_{u; v} = {}_c\bar{\textbf{K}}_{u; v} +  {}_p\Lambda_u 
\begin{bmatrix} 
\bar{\textbf{K}}_{a_{p-1};a_{q-1}} &  \bar{\textbf{K}}_{a_{p-1};a_q} \\
\bar{\textbf{K}}_{a_p;a_{q-1}} &  \bar{\textbf{K}}_{a_p;a_q}
\end{bmatrix}
{}_q\Lambda_v ^T,}
\end{equation}
where $ {}_c\bar{\textbf{K}}_{u; v}$ is provided in Equation (\ref{eq:kern_cond_cov}).\\

\textbf{\underline{Multivariate String GP Kernels}}

Univariate \textit{string GP kernels} may be combined for multivariate tasks following approaches such as hierarchical kernel learning (\cite{hkl}) or the additive kernels of \cite{DuvenaudNR2012}. Both approaches yield vastly more flexible kernels than multivariate kernels that use a specific norm (Euclidean or Mahalanobis) in the input space. Moreover, the special case where a product of univariate \textit{string GP kernels} is considered extends the Automatic Relevance Determination model of \cite{gp_intro}. In effect, not only can different input dimensions have different hyper-parameters, but each string in every input dimension may have its own hyper-parameters, allowing for \textit{Automatic Local Relevance Determination} (ALRD).
\section{Experiments}
\label{sct:exps}
\begin{figure}[ht]
\begin{center}
\vspace{-1em}
\centerline{\includegraphics[width=0.5\textwidth]{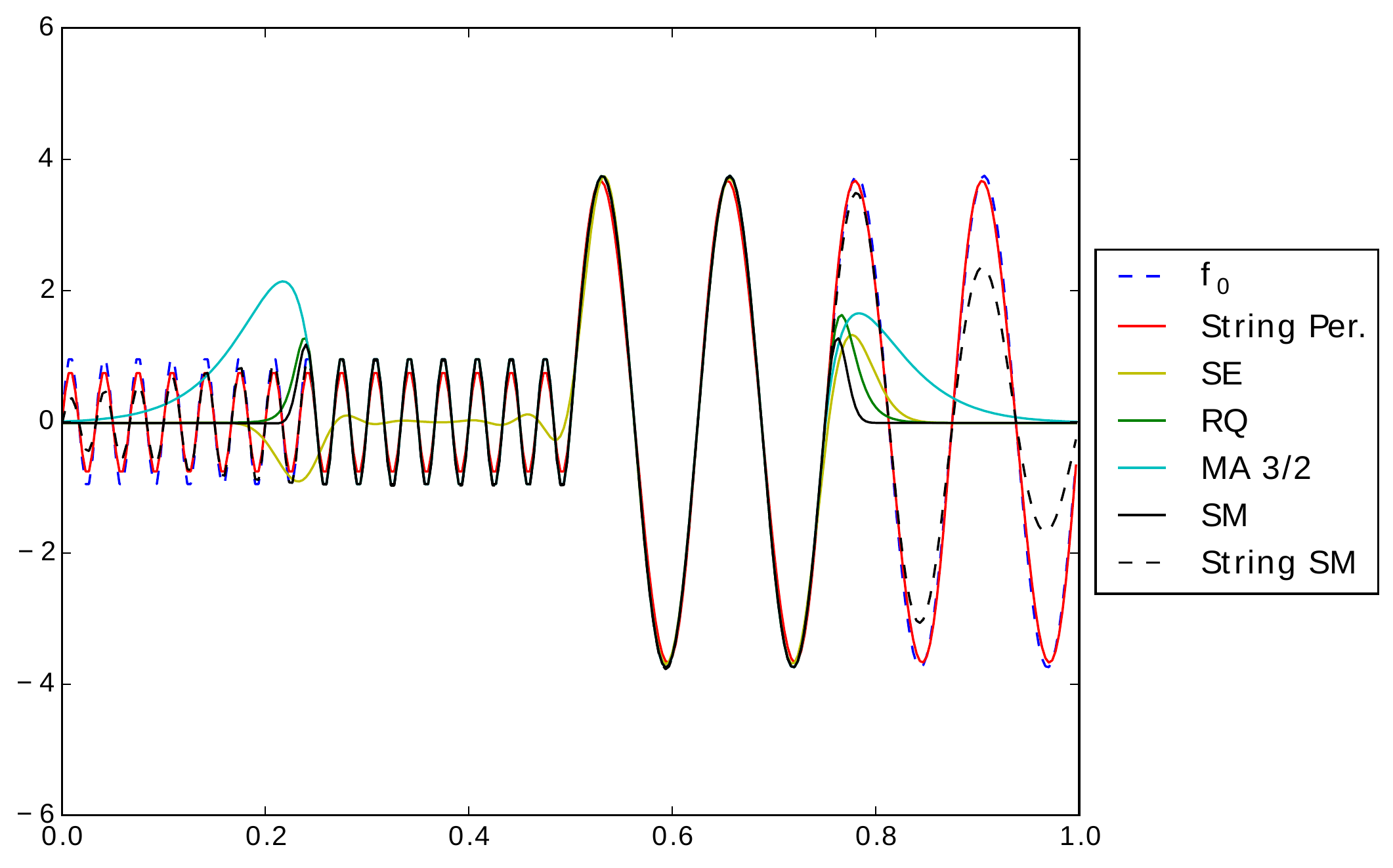}\includegraphics[width=0.5\textwidth]{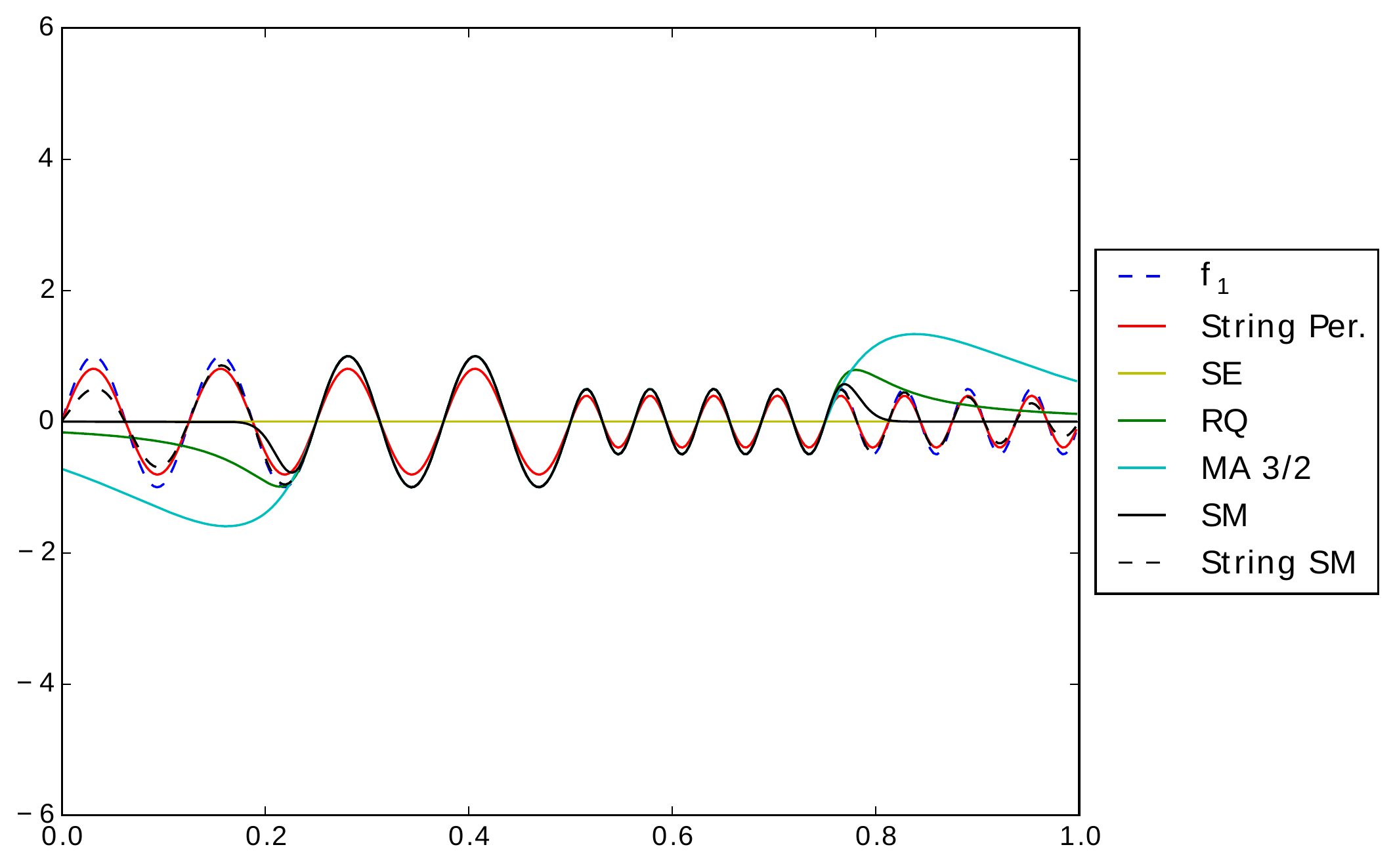}}
\vspace{-1em}
\caption{GP extrapolation of synthetic functions $f_0$ and $f_1$.}
\vspace{-1.5em}
\label{fig:synth_bench}
\end{center}
\end{figure}

\textbf{Synthetic data}:
In our first experiment, we highlight the limitation of standard GPs, in which a global covariance structure is postulated on the domain. This approach can result in unwanted global extrapolation of local patterns. We  show that this limitation is addressed by  \textit{string GP kernels}. We use $2$ toy regression problems with the following functions:
\begin{equation}
\label{eq:synth1d}
  \begin{array}{c c c}
f_0(t) = \left\{ 
  \begin{array}{l l}
    \sin(60\pi t) & ~ t \in [0, 0.5]\\
  \frac{15}{4}\sin(16\pi t) & ~ t \in ]0.5, 1]
  \end{array} \right., &
  f_1(t) = \left\{ 
  \begin{array}{l l}
    \sin(16\pi t) & ~ t \in [0, 0.5]\\
  \frac{1}{2}\sin(32\pi t) & ~ t \in ]0.5, 1]
  \end{array} \right..
    \end{array}
\end{equation}
$f_0$ (resp. $f_1$) undergoes a sharp (resp. mild) change in frequency and amplitude at $t=0.5$. We consider using their restrictions to $[0.25, 0.75]$ for training. We sample those restrictions with frequency $300$, and we would like to extrapolate the functions to the rest of their domain using GP regression (\cite{rasswill}). We compare \textit{string GP kernels} with string boundary times $\{0, 0.5, 1\}$ to popular and expressive kernels. We use as unconditional kernels for strings the periodic kernel of \cite{gp_intro} (String Periodic) and a spectral mixture kernel with a single spectral component per string (String Spec. Mix.). Figure \ref{fig:synth_bench} illustrates plots of the posterior means for each kernel used,  and Table \ref{table:synth_bench} compares predictive errors.
\begin{wraptable}{l}{0.5\linewidth}
\centering
\small
\vspace{-1em}
\begin{tabular}{@{}lll@{}}
\toprule
& \multicolumn{2}{c}{MAE} \\
\cmidrule{2-3} 
Kernel & $f_0$ & $f_1$    \\
\toprule
Squared Exponential & $0.89 \pm 2.01$ & $0.48 \pm 0.58$   \\
Rational Quadratic  & $0.71 \pm 2.18$ & $0.27 \pm 0.81$ \\
Matern 3/2 & $0.81 \pm 2.37$ & $0.55 \pm 1.45$  \\
Spec. Mix. (10 comp.) & $0.72 \pm 1.12$ & $0.20 \pm 0.30$  	\\ 
String Spec. Mix. & $0.23 \pm 0.84$ & $\textbf{0.06}\pm \textbf{0.21}$ \\
String Periodic & $\textbf{0.12} \pm \textbf{0.14}$ & $0.09 \pm 0.11$  \\
\bottomrule
\end{tabular}
\caption{Predictive mean absolute error $\pm$ 2 std on the GP extrapolations of $f_0$ and $f_1$.}
\vspace{-1.5em}
\label{table:synth_bench}
\end{wraptable}
Overall, it can be noted that \textit{string GP kernels} outperform competing kernels, including the expressive spectral mixture kernel $k_{\text{SM}}(r) = \sum_{a=1}^{A}  \omega_a^2 \exp(-2\pi^2r^2\sigma^2_a)\cos(2\pi r \mu_a)$ of \cite{wilson2013gaussian} with $A=10$ mixture components.\footnote{The sparse spectrum kernel of \cite{sparsespectrum} can be thought of as the special case $\sigma_a \ll 1$.} The comparison between the spectral mixture kernel and the string spectral mixture kernel is of particular interest, since spectral mixture kernels are dense in the family of stationary kernels, and thus can be regarded as a general purpose family to learn stationary kernels from the data. In our experiment, the string spectral mixture kernel with  a single mixture component per string significantly outperforms the spectral mixture kernel with 10 mixture components. This intuitively can be attributed to the fact that, regardless of the number of mixture components in the spectral mixture kernel, the learned kernel must account for both types of patterns present in each training dataset. Hence, each local extrapolation on each side of $0.5$ will attempt to make use of a mixture of both amplitudes and both frequencies evidenced in the training datasets, and will struggle to recover the true local sine function. It should be expected that the performance of the spectral mixture kernel in this experiment will not improve drastically as the number of mixture components increases. However, under a string GP prior, the left and right hand side strings are independent conditional on the (unknown) boundary conditions. Therefore, the training dataset on $[0.25, 0.5]$ influences the hyper-parameters of the string to the right of $0.5$ only to the extent that both strings should agree on the value of the latent function and its derivative at $0.5$. To see why this is a weaker condition, we consider the family of pair of functions: $(\alpha \omega_1 \sin(\omega_2 t), ~  \alpha \omega_2 \sin(\omega_1 t)), ~ \omega_i = 2\pi k_i, k_i \in \mathbb{N}, \alpha \in \mathbb{R}.$
Such functions always have the same value and derivative at $0.5$, regardless of their frequencies, they are plausible GP paths under a spectral mixture kernel with one single mixture component ($\mu_a = k_i$ and $\sigma_a \ll 1$), and can be exactly reproduced with the periodic kernel of \cite{gp_intro}. As such, it is not surprising that extrapolation under a string spectral mixture kernel should perform well, and that a string periodic kernel should recover the function almost perfectly. 

%
%
\begin{figure}[ht!]
\begin{center}
\centerline{\includegraphics[width=0.5\textwidth]{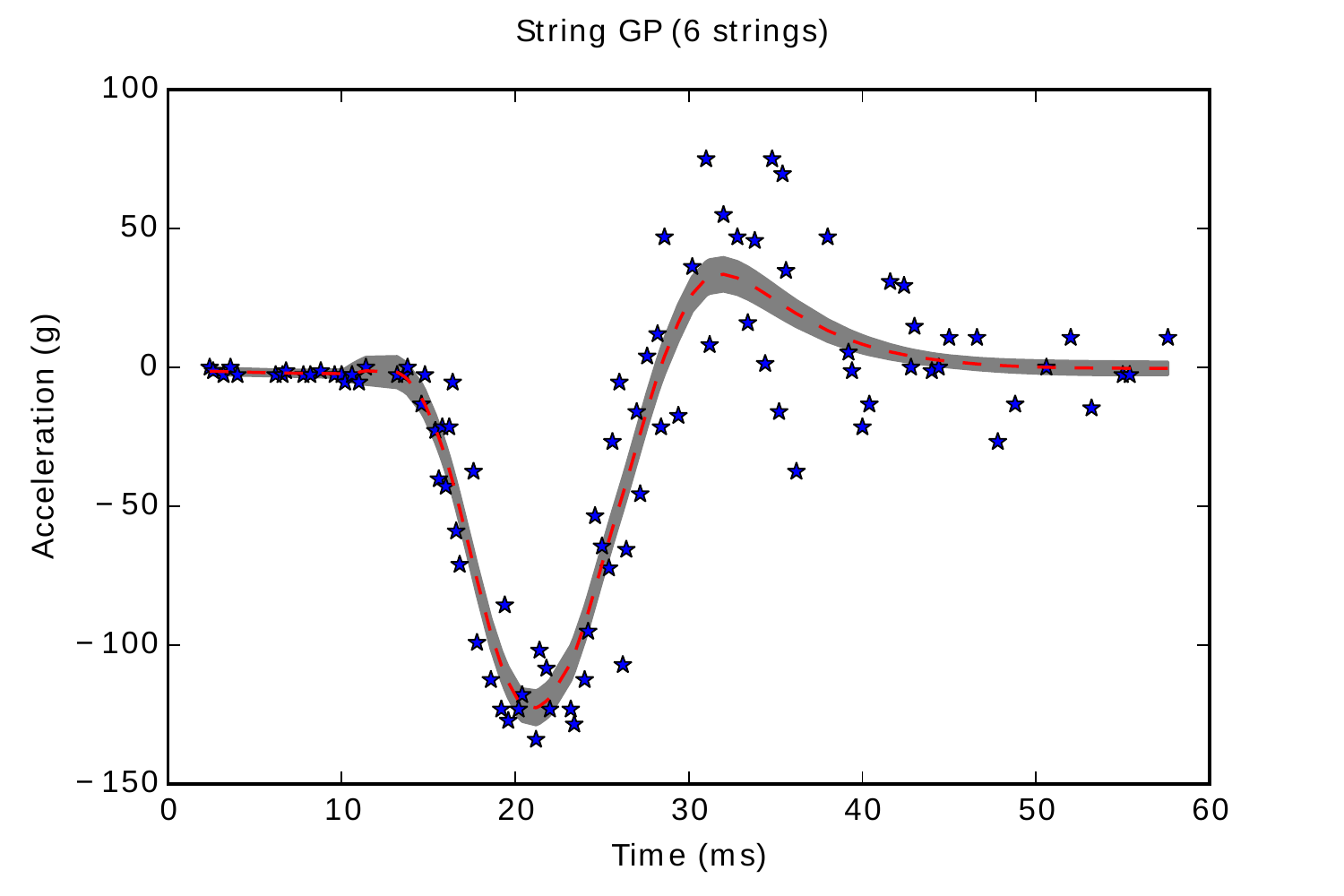}\includegraphics[width=0.5\textwidth]{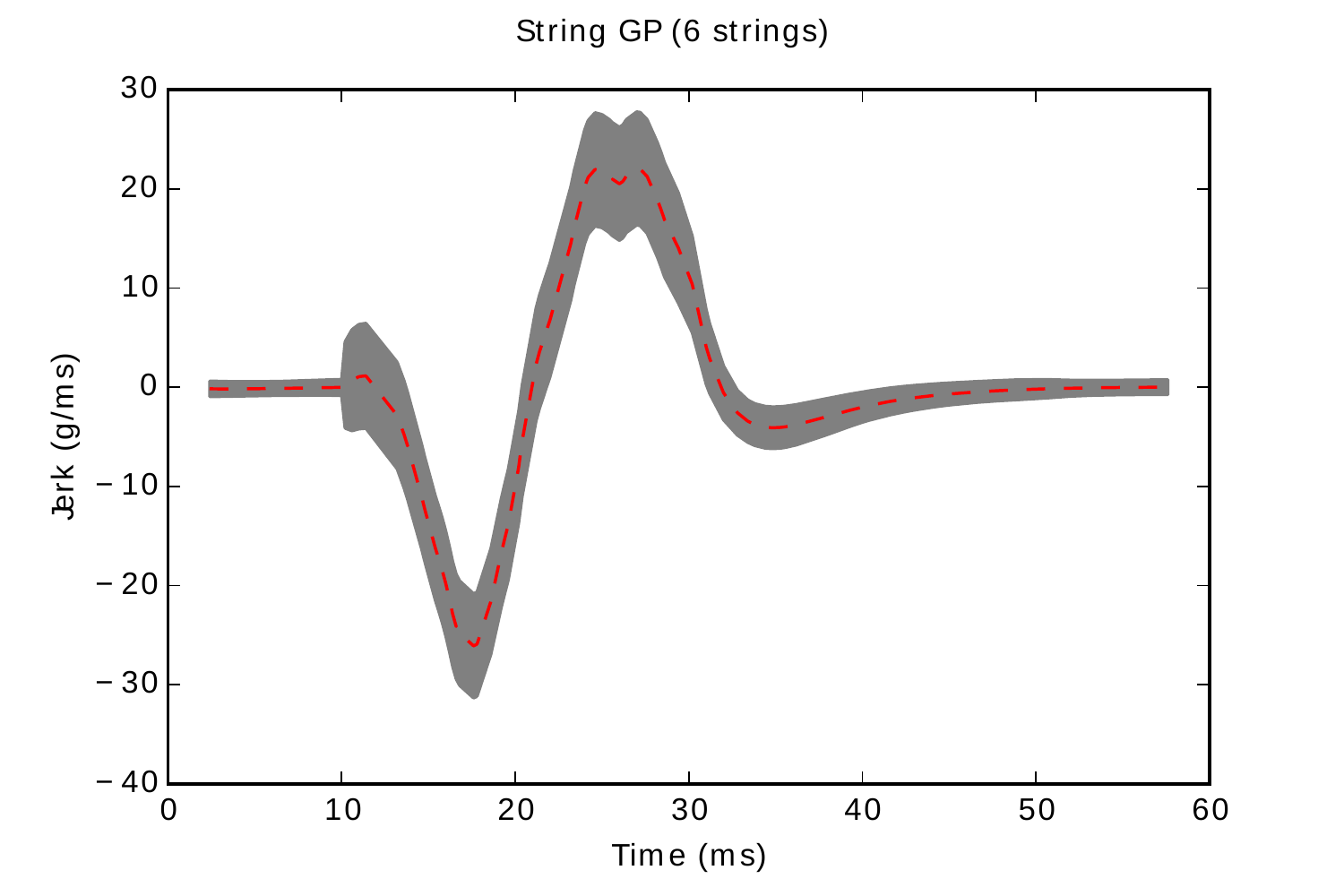}}
\vspace{-1em}
\caption{Posterior mean $\pm$ 2 predictive standard deviations on the motorcycle dataset (\cite{silverman}), under a Matern 3/2 derivative string GP prior with 6 equal length string intervals.}
\vspace{-1.5em}
\label{fig:motorcycle_jerk}
\end{center}
\end{figure} 

\textbf{Motorcycle data:}
The objective of this experiment is three-fold. Firstly, we show on the well studied motorcycle dataset of \cite{silverman} that using guestimates for the partition of the domain, along with string GPs, outperforms the standard GP model. Secondly, we show that our approach considerably outperforms the alternatives proposed by \cite{kim, gramacy}, that consist of dividing the training dataset in disjoint subsets and performing independent training; this confirms the importance of the collaboration between experts. Thirdly we illustrate learning the derivative of the function from noisy measurements. The observations consist of accelerometer readings taken through time in an experiment on the efficacy of crash helmets. We consider inferring the latent acceleration curve using GP regression, with an heteroskedastic Gaussian noise model, allowing the noise variance to be constant within a string interval and different between string intervals. We consider the two sets of boundary times $\{0, 15, 30, 45, 60\}$ and $\{0, 10, 20, 30, 40, 50, 60\}$. We ran $50$ independent random experiments, leaving out $5$ points selected uniformly at random from the dataset for prediction, the rest being used for training. The models we compared include a vanilla GP, for each of the above set of boundary times the corresponding string GP, and mixtures of independent GP experts. The Matern 3/2 kernel was used as base throughout. The results are reported in Table \ref{table:motorcycle_bench}.

\begin{wraptable}{l}{0.71\linewidth}
\centering
\small
\vspace{-1em}
\begin{tabular}{@{}lrrrcrrr@{}}\toprule
& Training & \multicolumn{4}{c}{Prediction} \\
\cmidrule{3-6} \cmidrule{2-2} 
& 						Log lik.	& Log lik. & 			MAE & 		Avg std& 		RR std    \\ \midrule
Vanilla GP&					-420.80 	& -22.71 & 			16.24& 		3.10 & 		0.23   \\

String GP (4 strings)&			-376.51	& \textbf{-20.54} & 	16.21 & 		\textbf{2.02} & 	1.26 \\ 
String GP (6 strings)&			-374.62 	& -20.58 &			\textbf{15.83}& 	2.14 & 		1.03  \\

Mix. of 4 GPs & 				-367.23 	& -29.31 & 			20.13 & 		7.58 & 		\textbf{1.86}  \\
Mix. of 6 GPs & 				-456.04 	& -39.11 & 			28.86  & 		17.53 & 		1.12  \\ 

Regul. Mix. of 4 GPs &			 -314.55	 & -22.80 &			17.41  & 		4.27 &		1.34 \\
Regul. Mix. of 6 GPs & 		\textbf{-273.68}	 & -22.73 &			18.32 & 		3.78 &		1.25  \\
\bottomrule
\end{tabular}
\caption{Performance comparison on the motorcycle dataset (\cite{silverman}). RR std is the average relative range (i.e. (max - min)/avg) of the posterior std.}
\vspace{-1em}
\label{table:motorcycle_bench}
\end{wraptable}

It can be seen from Table \ref{table:motorcycle_bench} that \textit{string GPs} outperform the vanilla GP and mixtures of GPs in terms of average out-of-sample predictive log likelihood of the (noisy) left out accelerations, regardless of the set of boundary times used. More importantly, we note that this also held true individually for all of the $50$ random experiments we ran. Moreover, string GPs yielded more certain predictions (Avg std) for the values of the latent function at test inputs than competing models. Furthermore, the heteroskedastic noise model, which fits naturally within the string GP framework, allows a wider range of posterior standard deviation (RR std) than the vanilla GP. Predictive certainty is indeed higher under string GP priors at the beginning and end of the experiment, which is intuitive as the magnitude of the noise in those parts is smaller. As for mixtures of independent GPs (\cite{kim, gramacy}), they significantly and consistently underperform both vanilla and string GPs out-of-sample, primarily because the latent acceleration is expected to be continuous. We added an $L^2$ penalty on the input length scale to the marginal likelihood in order to mitigate the discontinuity of the learned posterior means. Although the results are improved, their predictive performances remain worse than string GPs', and they are prone to over-fitting. We also learned the derivative of the latent acceleration with respect to time (Figure \ref{fig:motorcycle_jerk} right), purely from noisy acceleration measurements using the joint law of a string GP and its derivative (Theorem \ref{theo:sgp}).

\begin{figure}
\centering
\begin{minipage}[t]{0.5\textwidth}
	\small
	\centering
	\vspace{0pt}
	\begin{tabular}{@{}lrrrrl@{}}\toprule
	& Training & \multicolumn{3}{c}{Prediction} \\
	\cmidrule{3-6} \cmidrule{2-2} 
	& 				Log lik.		& Log lik. & 			MAE& 			Avg std \\ \midrule
	ARD Poly &		-452.32 		& -39.06 & 			0.61&			0.62 \\ 
	ALRD Poly &		-423.68 		& -38.80 & 			0.65&			0.56 \\  [-0.1ex] \hdashline  \\ [-1.5ex]
	 
	ARD SE &			-452.32 		& -39.06 & 			0.61&			0.61 \\ 
	ALRD SE &		-423.24 		& -38.64 & 			0.66&			0.56 \\  [-0.1ex] \hdashline  \\ [-1.5ex]
	
	ARD RQ &		-452.17 		& -38.37 & 			0.59 &			0.60 \\ 
	ALRD RQ &		-426.90		&\textbf{ -28.88} & 	\textbf{0.44} &	\textbf{0.25} \\  [-0.1ex] \hdashline  \\ [-1.5ex] 
	
	ARD MA 3/2 &	 	-452.37 		& -33.49 & 			0.50&		 	0.50  \\  
	ALRD MA 3/2 &	-423.14 		& -33.09 & 			0.55&			0.40 \\ [-0.1ex] \hdashline  \\ [-1.5ex]
	 
	ARD MA 5/2 &		-452.09 		& -36.24 & 			0.55&			0.52 \\
	ALRD MA 5/2 &	\textbf{-419.62} 	& -35.28 & 			0.59&			0.44 \\  
	\bottomrule
	\end{tabular}
	\captionof{table}{ Automatic Local Relevance Determination using \textit{string GP kernels} on the global air temperature anomalies dataset.}
	\label{table:air_temp}
\end{minipage}\hfill
\begin{minipage}[t]{0.44\textwidth}
	\centering
	\vspace{-0.2em}
	\includegraphics[width=\textwidth]{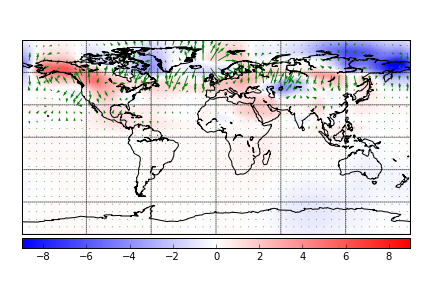}
	\vspace{-1.5em}
	\captionof{figure}{Posterior mean and gradient vector field of the global air temperature anomalies of \cite{wood_temp} (${^\circ}$C) in December 1993, under a string GP Matern 3/2 kernel.}
	\label{fig:air_temp}
\end{minipage}
	\vspace{-1em}
\end{figure}

\textbf{Global air temperature anomalies:} Next, we illustrate \textit{Automatic Local Relevance Determination} by comparing popular ARD kernels with products of univariate \textit{string GP kernels}. We also illustrate that the gradient of the latent function can be learned jointly with the latent function (see appendix). The experiment is based on the well studied temperature anomalies dataset of \cite{wood_temp}. The dataset consists of monthly readings of air temperature anomalies at various points on the globe in December 1993. The authors defined air temperature anomaly as the deviation of a monthly temperature at a given location from the average over the period 1950-1979 of the monthly temperatures at the same location. There were $445$ readings in December 1993. We ran 50 experiments, each using 50 points selected uniformly at random for testing and using the rest for training. We considered several popular kernels; a product of univariate second order polynomial kernels (i.e. $k(x, y)=\sigma^2(xy+c)^2, ~\sigma, c >0$), the ARD squared exponential kernel, the ARD rational quadratic  kernel, and the ARD Matern 3/2 and 5/2 kernels. We compared each of the above kernels with a product of univariate \textit{string GP kernels} that has unconditional kernels in the same family and with boundary times $\{-90, 0, 90\}$ in the latitude dimension and $\{0, 180, 360\}$ in the longitude dimension. Average training and predictive results are summarized in Table \ref{table:air_temp}. We note that \textit{string GP kernels} (ARLD) have higher predictive log likelihood and yield more certain predictions than their `non-string' counterparts, despite the simplicity of our choice of boundary times. Although ARD kernels sometimes yielded lower mean absolute errors, the difference in those cases was never in excess of $0.05^\circ$C, which corresponds to just $2.5\%$ of the sample standard deviation of anomalies ($1.99 ~^\circ$C) and about $6\%$ of the learned noise standard deviation (ranging from $0.82^\circ$C to $0.98^\circ$C). Figure \ref{fig:air_temp} illustrates a map of the globe with the posterior mean of the latent anomalies and the posterior mean gradient vector field.

\section{Discussion}
\label{sct:dsc}
In this paper we introduce a new class of nonstationary kernels we refer to as \textit{string GP kernels}, that allow learning heterogeneous local patterns in the data, while ensuring global smoothness. We demonstrate the need for our approach on synthetic data, and we show that our approach outperforms competing alternatives on well studied real-life regression problems. 

\textbf{Domain partition:} As illustrated by the last two experiments, in many applications, naively partitioning the domain might considerably improve on stationary kernels. The accuracy may be further improved by considering uniform partitions of the domain and learning the partition size by cross-validation using predictive likelihood or root mean square error as out-of-sample performance metric. Alternatively, the number and positions of boundary times may be jointly learned in a fully Bayesian setting by putting a point process prior on boundary times, and sampling from both the posterior intensity and the positions of the points using MCMC.

\textbf{Future work:} Although the focus of this paper has been on improving predictive accuracy, the conditional independence between local experts may be exploited to develop exact inference methods that scale an order of magnitude better than the standard approach.
\newpage
\bibliography{arxiv_sgpk}
\newpage
\clearpage

 \toptitlebar
\begin{center}{\centering \LARGE\bfseries Supplementary Material for String GP Kernels}\end{center}
 \bottomtitlebar

\begin{appendices}
\renewcommand\thesection{Appendix \Alph{section}}
\section{Additional Results}
\subsection{Motorcycle dataset}
\begin{figure}[ht!]
\begin{center}
\centerline{\includegraphics[width=0.45\textwidth]{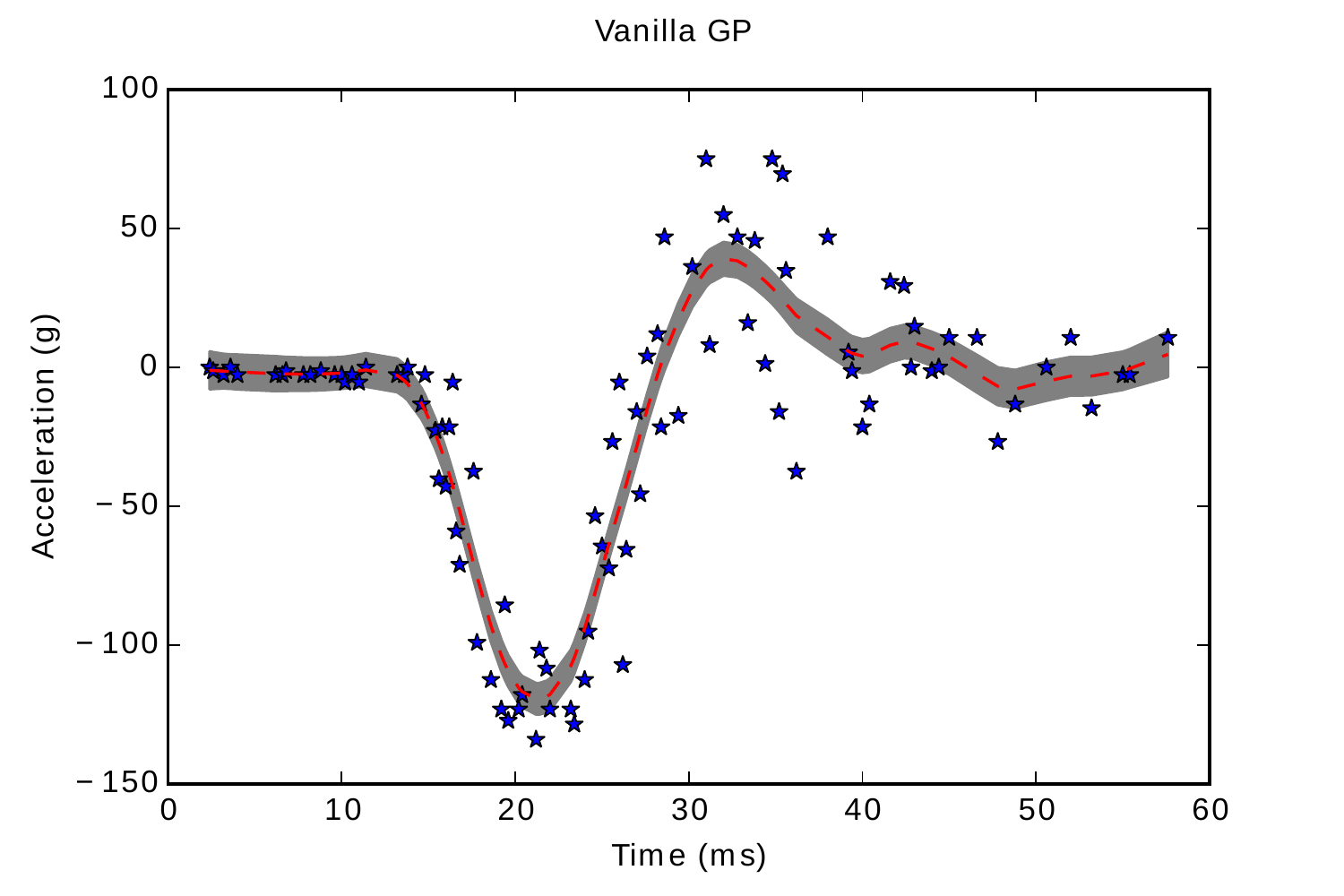}}
\centerline{\includegraphics[width=0.45\textwidth]{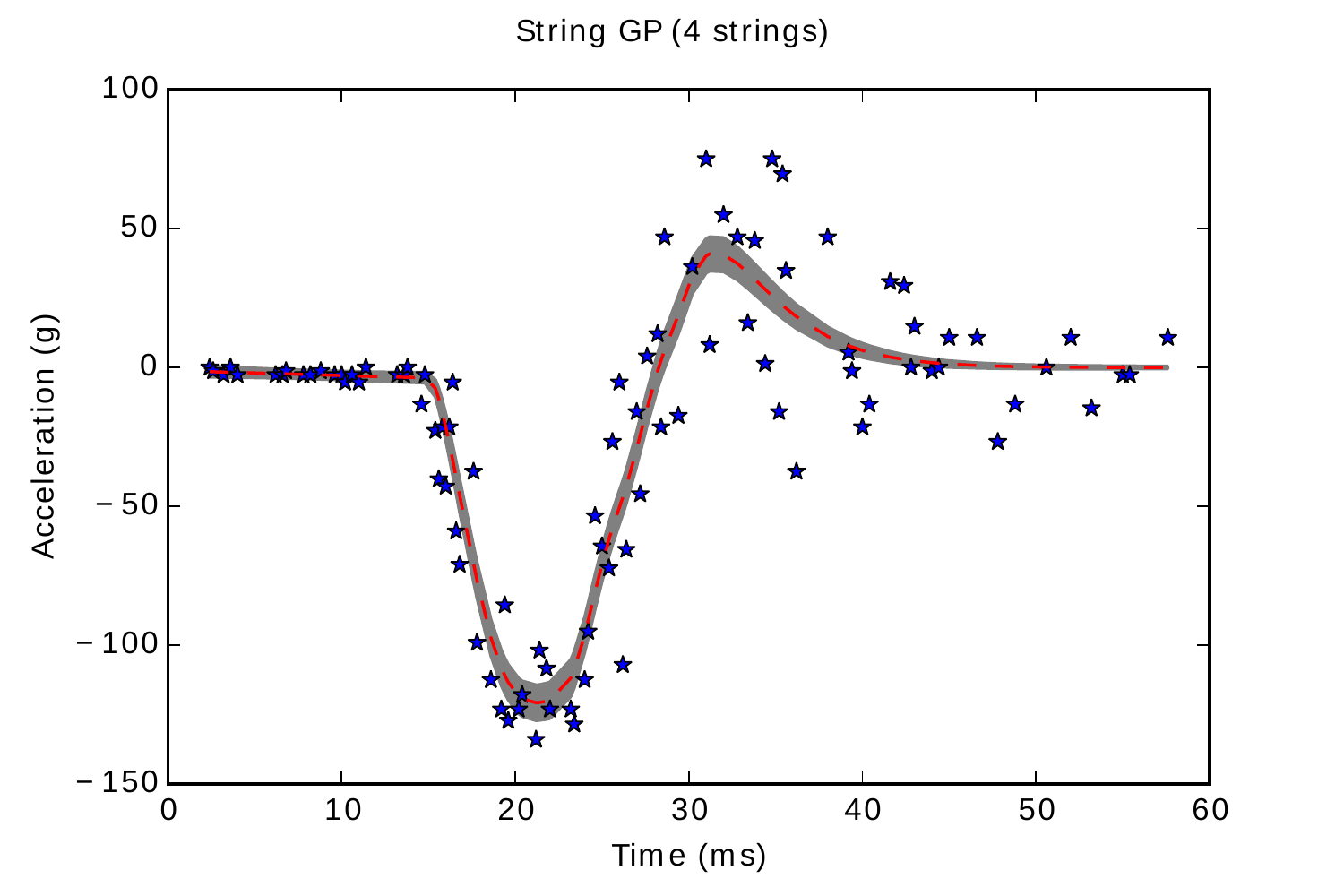}\includegraphics[width=0.45\textwidth]{motocycle_sgp_6}}
\centerline{\includegraphics[width=0.45\textwidth]{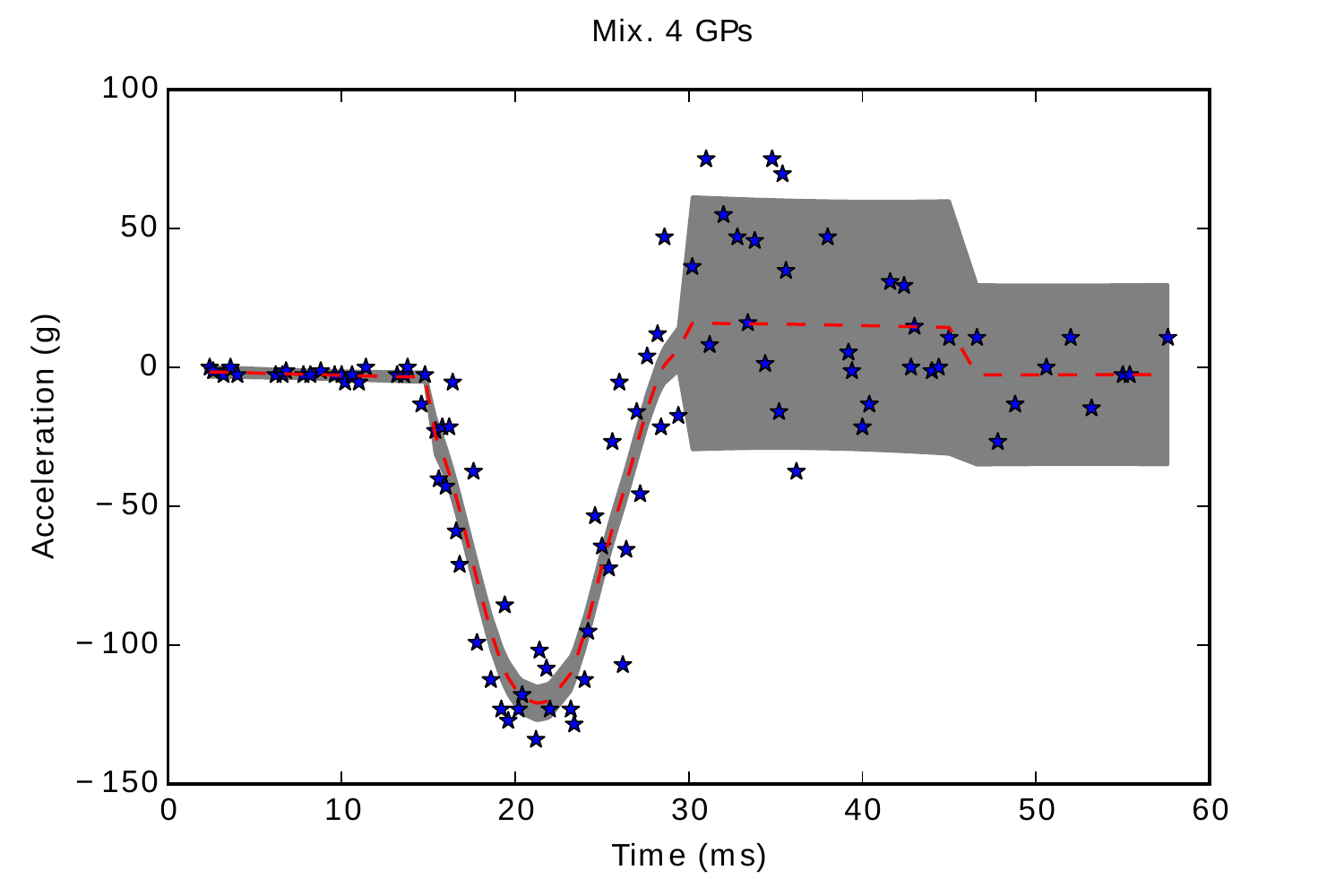}\includegraphics[width=0.45\textwidth]{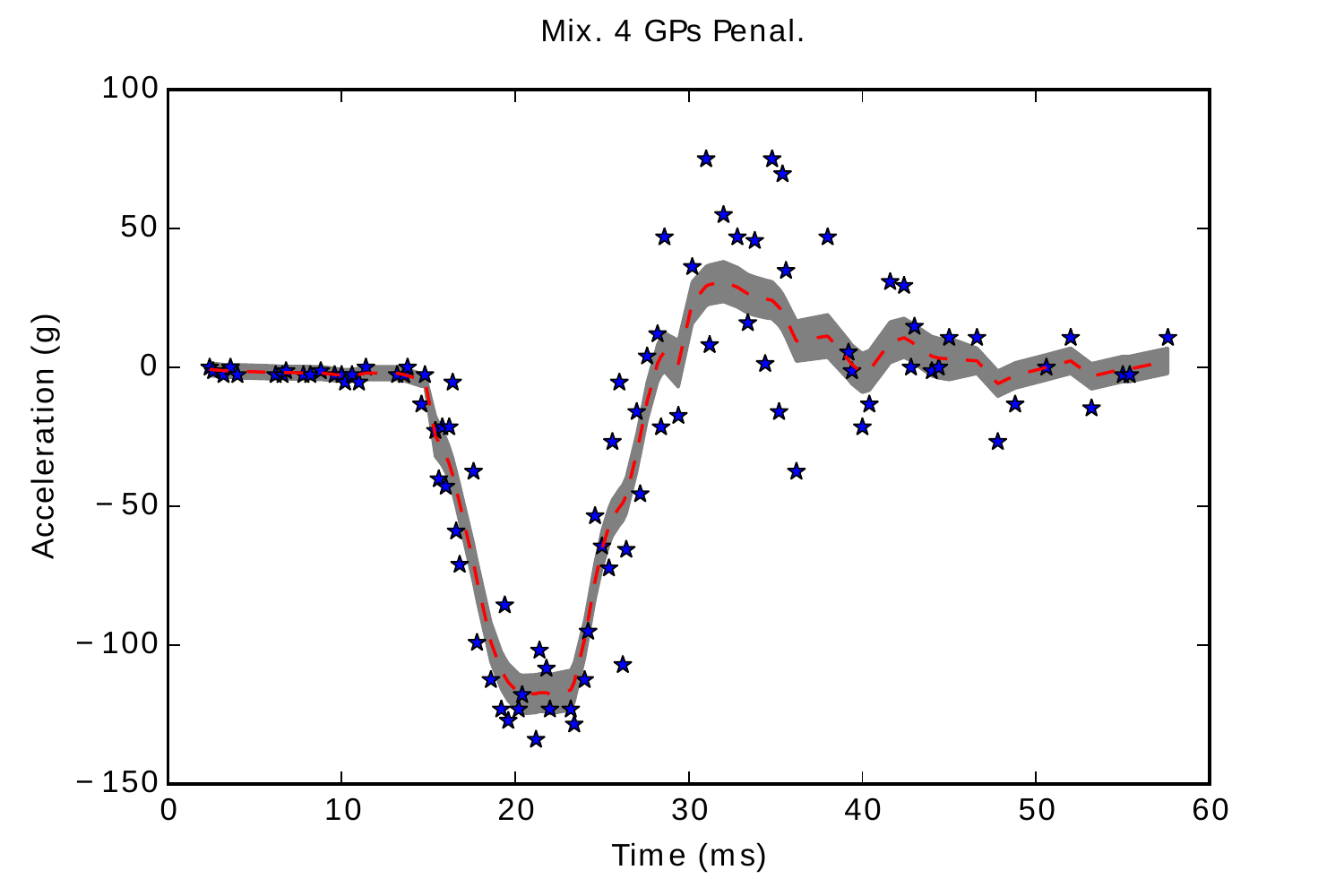}}
\centerline{\includegraphics[width=0.45\textwidth]{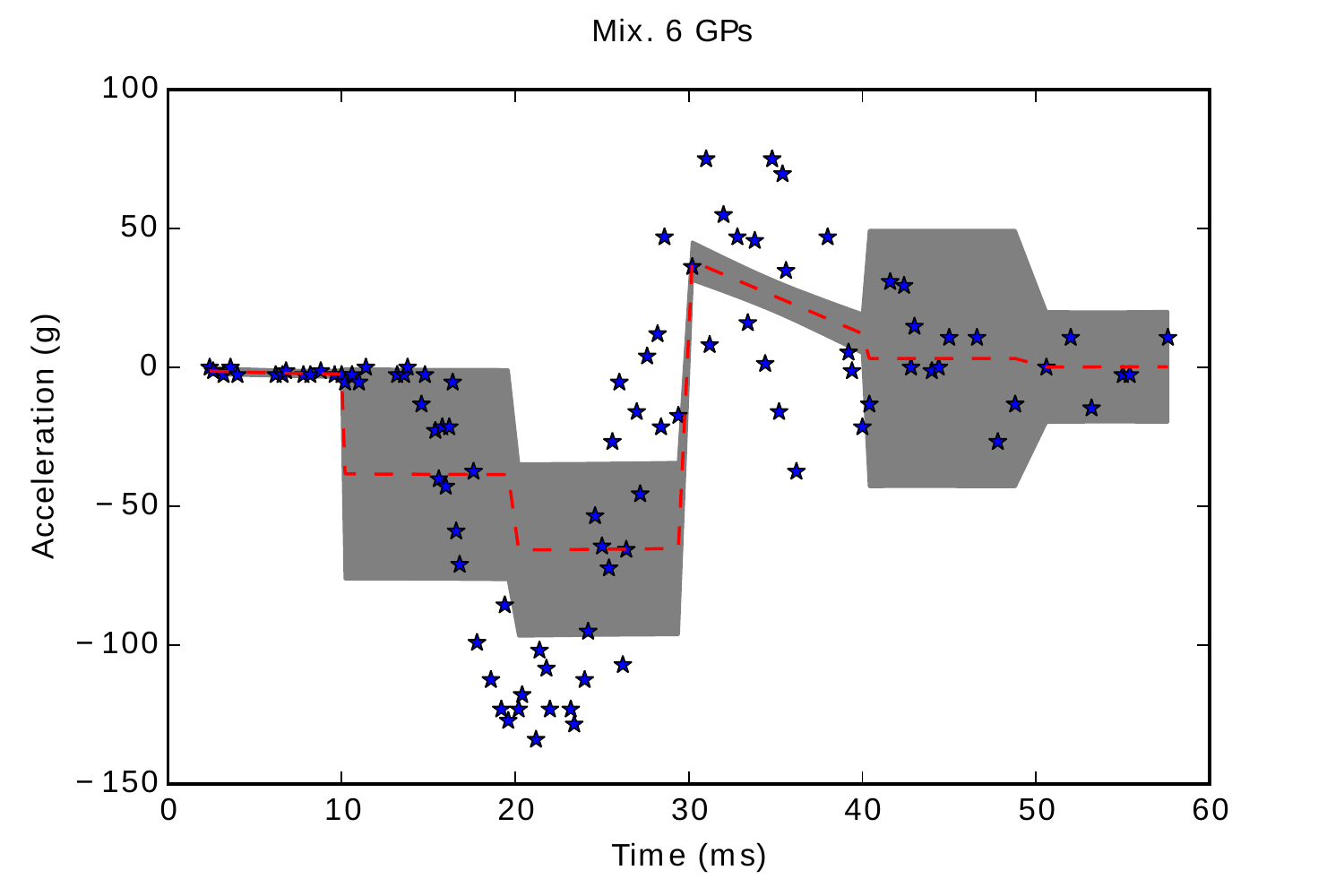}\includegraphics[width=0.45\textwidth]{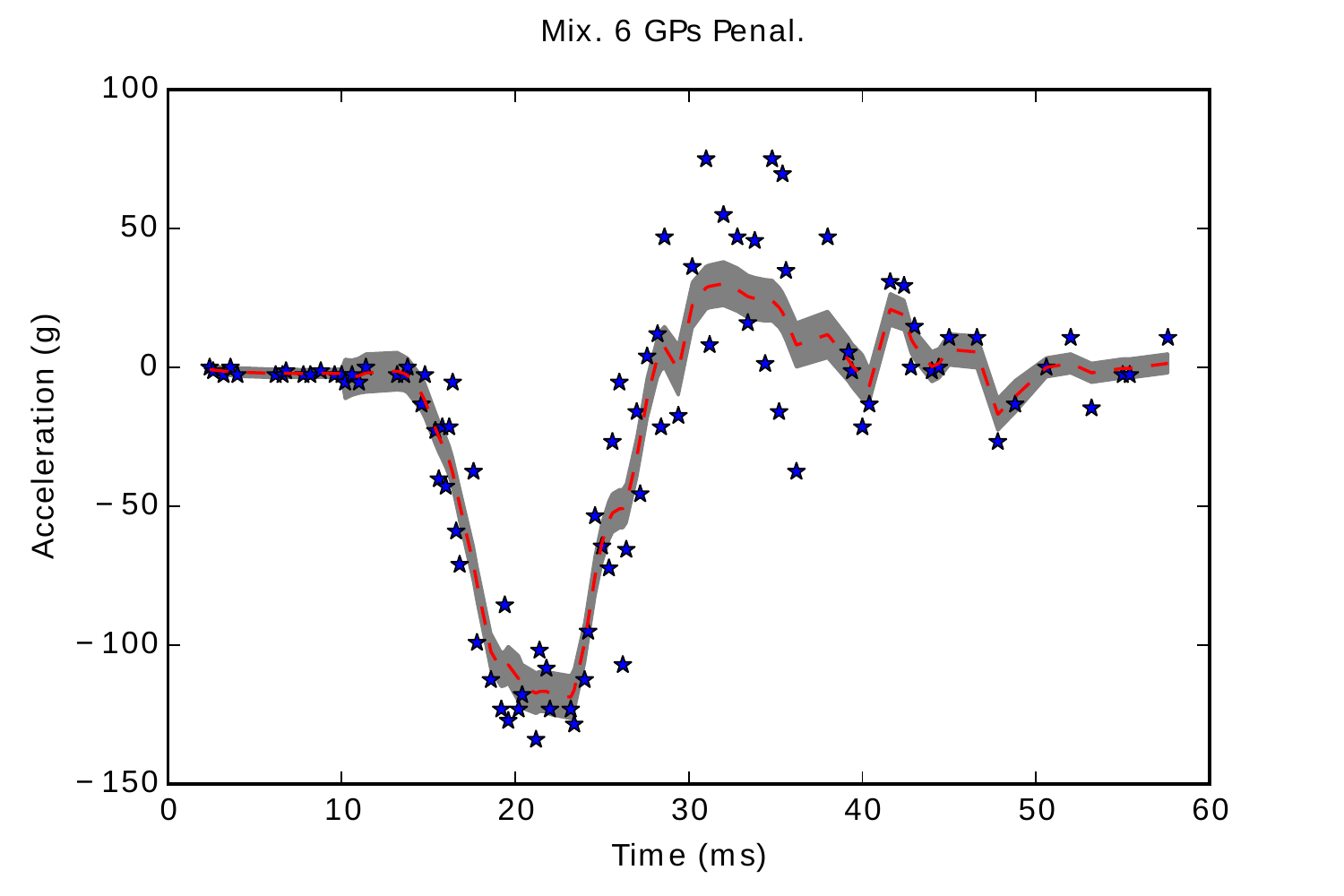}}
\caption{Posterior mean $\pm$ 2 predictive std on the motorcycle dataset \cite{silverman} under a vanilla GP, string GPs, and mixtures of GP experts (with and without prior on the input scale).}
\label{fig:motorcycle}
\end{center}
\end{figure}

Figure \ref{fig:motorcycle} illustrates posterior means and  $\pm$ 2 standard deviations confidence bands for the motorcycle experiment.

\subsection{Additional Experiments}
\begin{figure}[ht]
\begin{center}
\centerline{\includegraphics[width=0.45\textwidth]{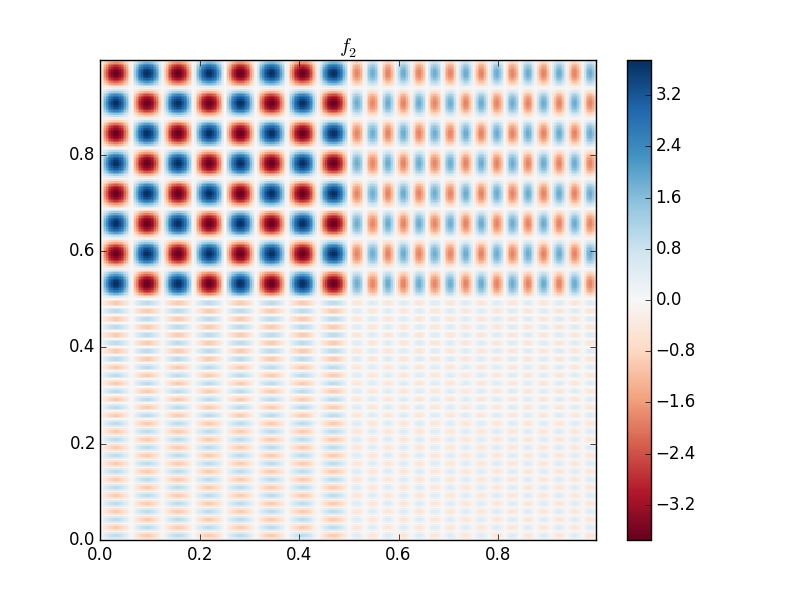} \includegraphics[width=0.45\textwidth]{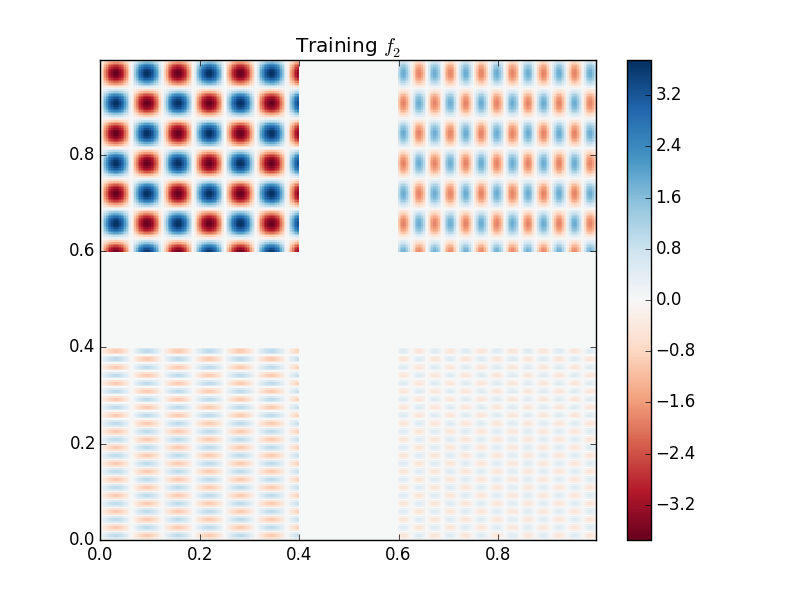}} 
\centerline{\includegraphics[width=0.45\textwidth]{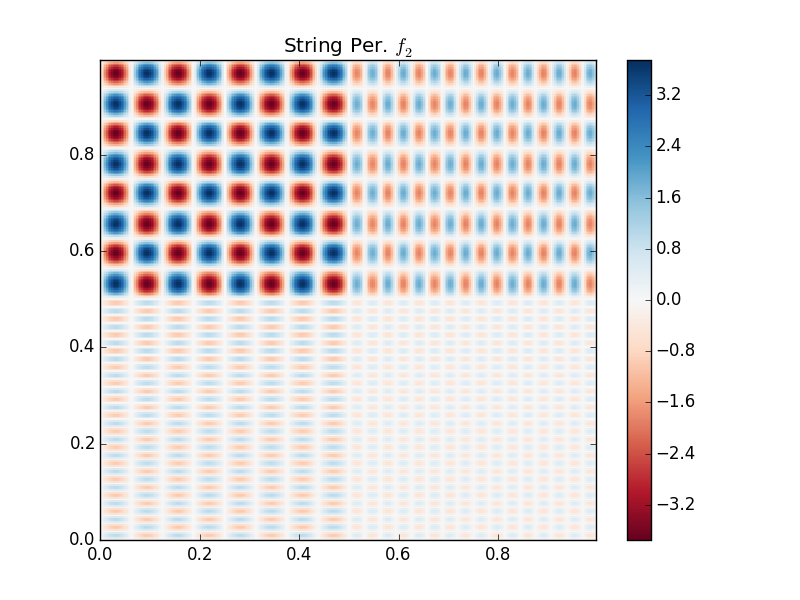} \includegraphics[width=0.45\textwidth]{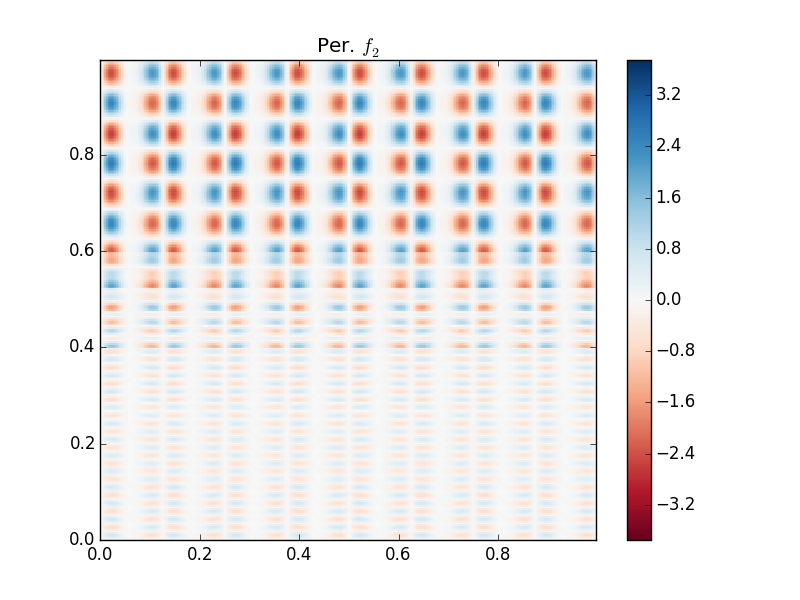}}
\centerline{\includegraphics[width=0.45\textwidth]{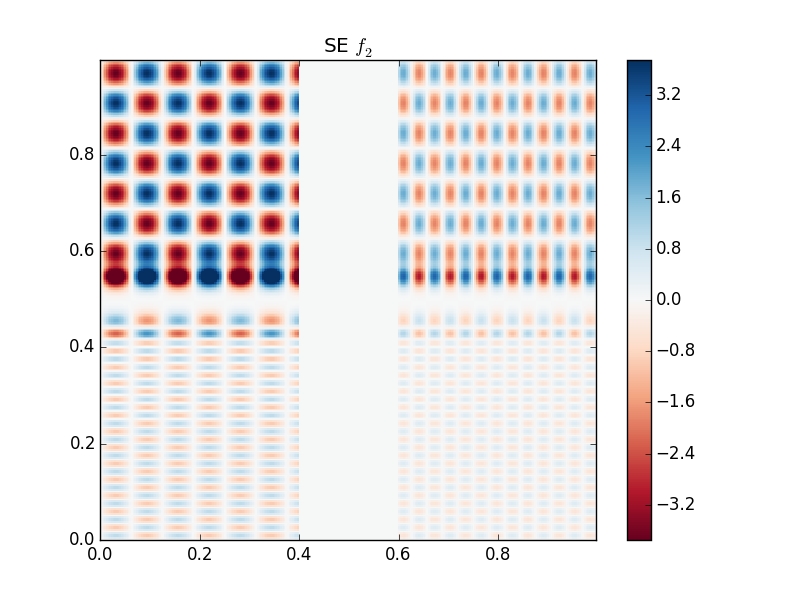} \includegraphics[width=0.45\textwidth]{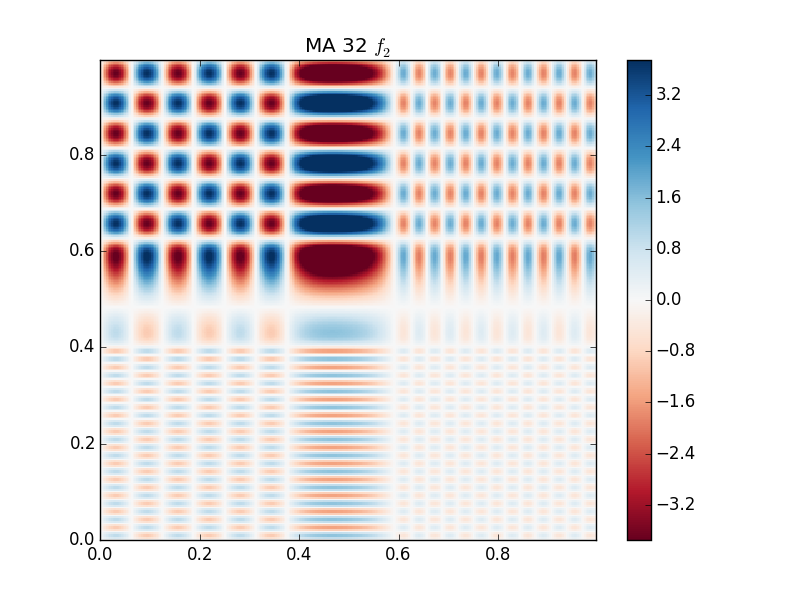}}
\centerline{\includegraphics[width=0.45\textwidth]{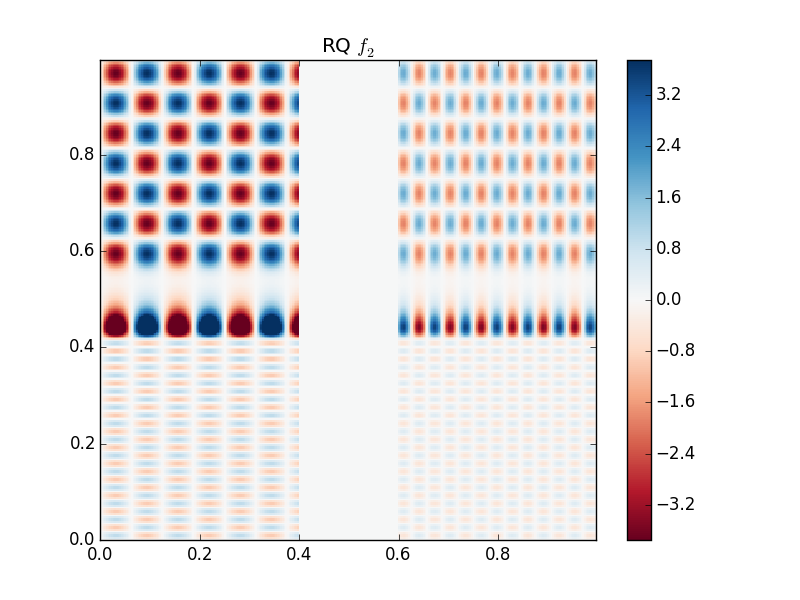} \includegraphics[width=0.45\textwidth]{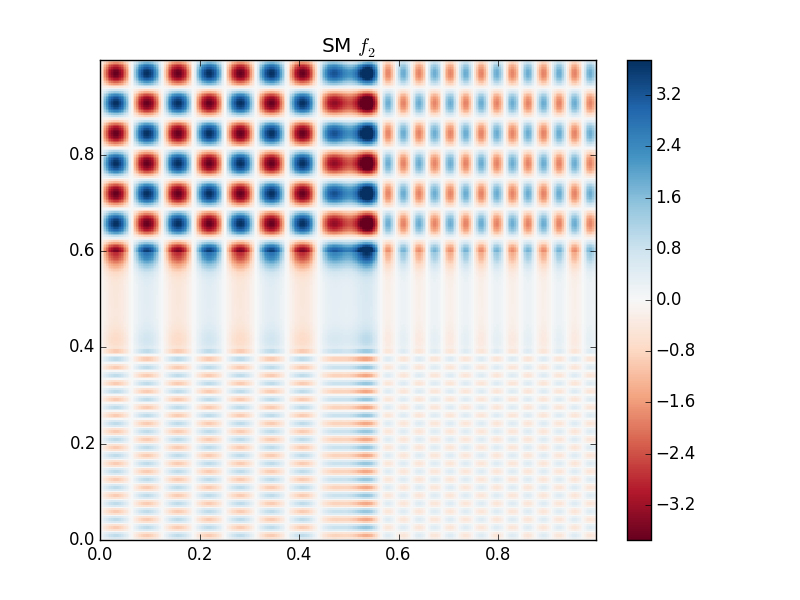}}
\caption{Comparison of kernels in the GP interpolation of $f_2$.}
\label{fig:pred_f2}
\end{center}
\end{figure}

\begin{figure}[ht]
\begin{center}
\centerline{\includegraphics[width=0.45\textwidth]{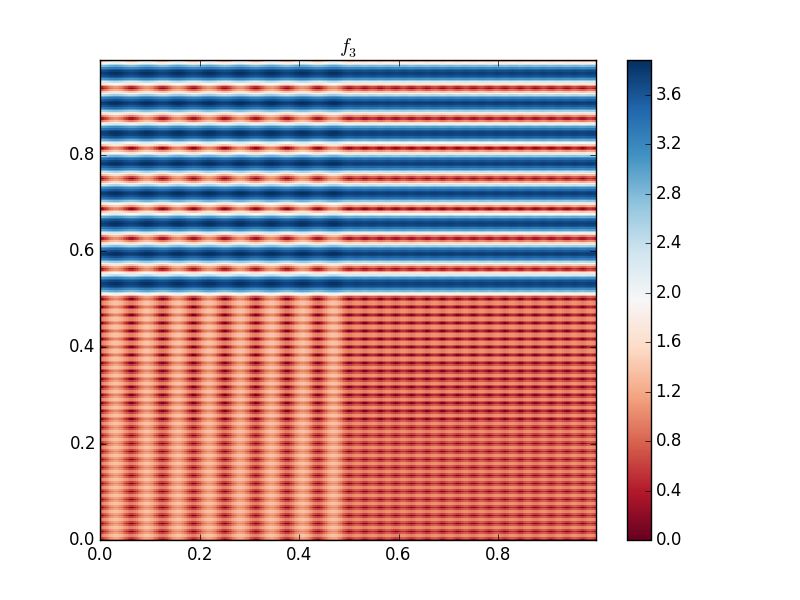} \includegraphics[width=0.45\textwidth]{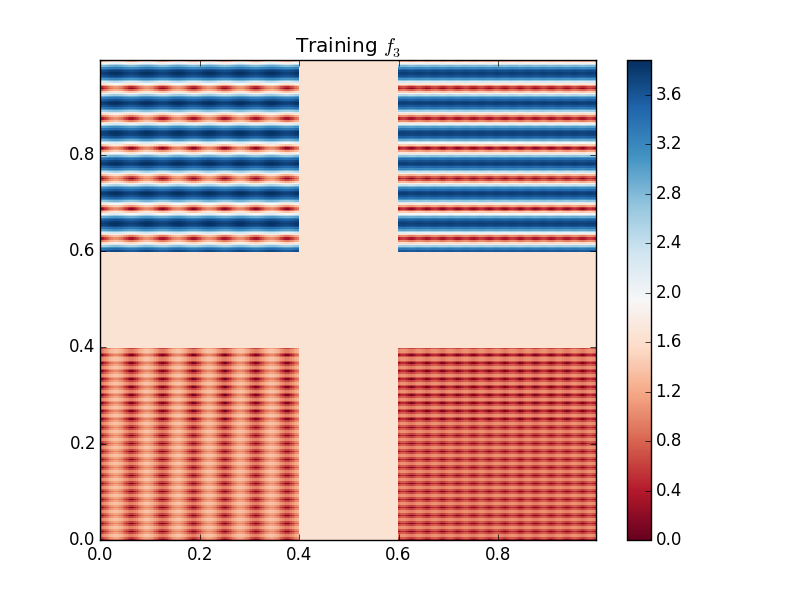}} 
\centerline{\includegraphics[width=0.45\textwidth]{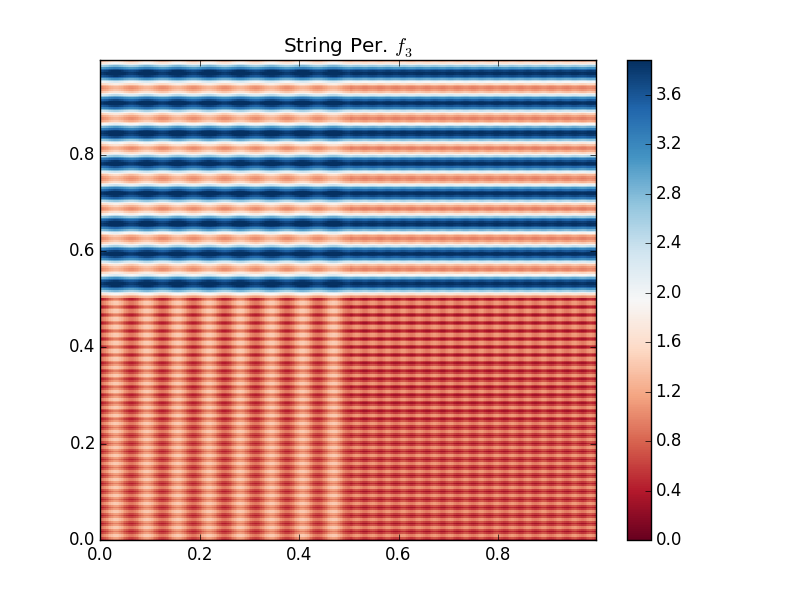} \includegraphics[width=0.45\textwidth]{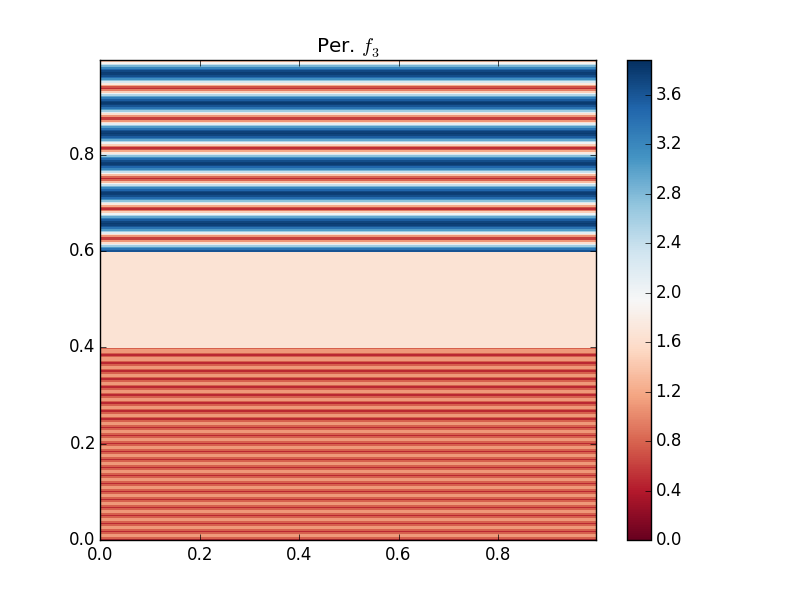}}
\centerline{\includegraphics[width=0.45\textwidth]{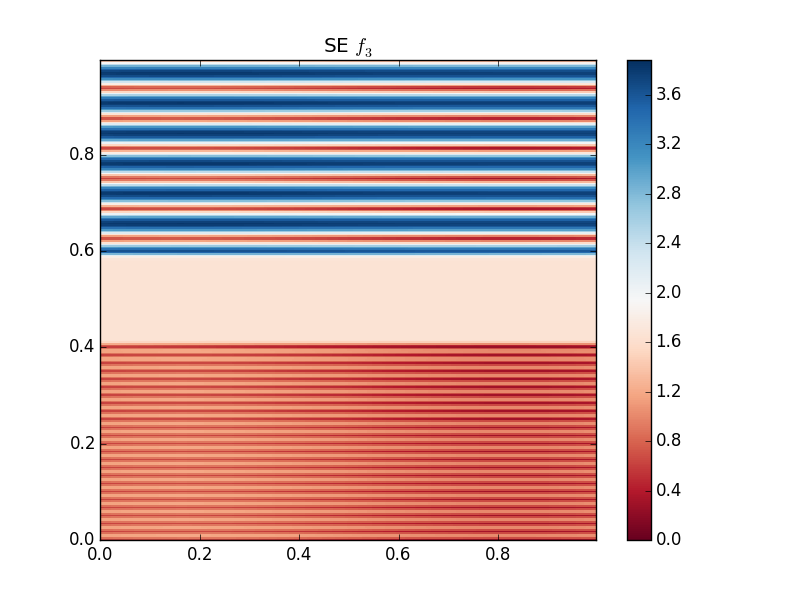} \includegraphics[width=0.45\textwidth]{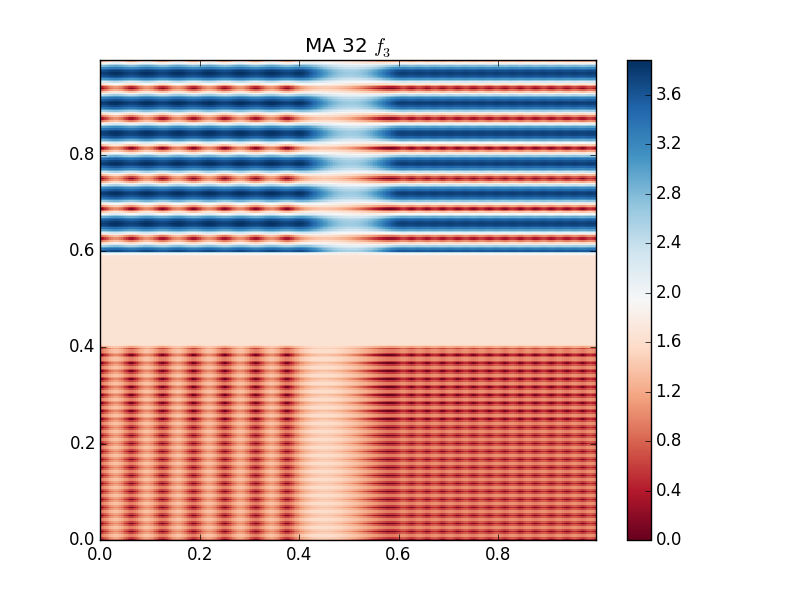}}
\centerline{\includegraphics[width=0.45\textwidth]{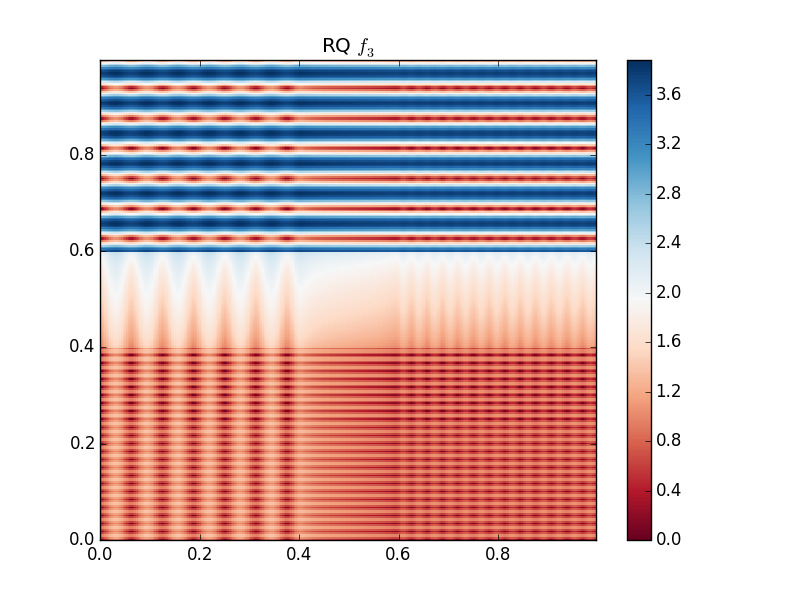} \includegraphics[width=0.45\textwidth]{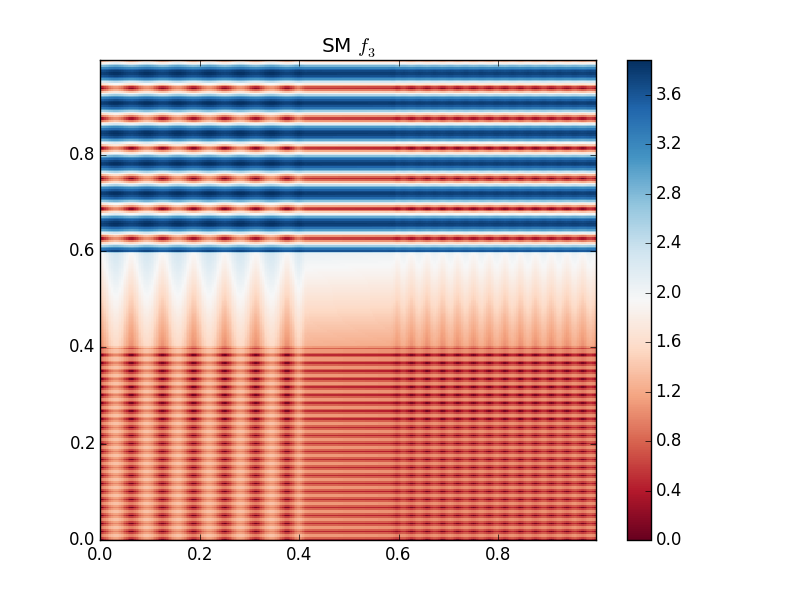}}
\caption{Comparison of kernels in the GP interpolation of $f_3$.}
\label{fig:pred_f3}
\end{center}
\end{figure}

In the first experiment we have demonstrated that \textit{string GP kernels}  outperform competing kernels on univariate extrapolation tasks in the presence of local patterns. This  additional experiment aims at reinforcing this message by considering two bivariate interpolation problems exhibiting local patterns, and with inputs forming a grid. We introduce two functions $f_2$ and $f_3$:
\begin{equation}
\forall u, v \in [0, 1], ~f_2(u, v) = f_0(u)f_1(v) ~~\text{and}~~ f_3(u, v) = \sqrt{f_0^2(u) + f_1^2(v)},
\end{equation}
where $f_0$ and $f_1$ are the functions of our univariate synthetic experiment. For training, we use the restrictions of $f_2$ and $f_3$ to $[0, 0.4] \cup [0.6, 1] \times [0, 0.4] \cup [0.6, 1]$, which we evaluate on a uniform grid with mesh size $1/300$ (72000 training points). We consider recovering the two functions on $[0,1] \times [0, 1]$  using GP regression with i.i.d. Gaussian noise terms (90000 test inputs), and compare a product of \textit{string GP kernels} with competing alternatives. We refer to \cite{wilson2013gaussian}, for tricks to leverage the Kronecker structure of the resulting covariance matrix to yield inference in $\mathcal{O}(N^\frac{3}{2})$ computational time and $\mathcal{O}(N)$ memory requirement. We compare a product of string periodic kernels (String Per.), a product of periodic kernels introduced by \cite{gp_intro} (Per.), a product of squared exponential kernels (SE), a product of Matern 3/2 kernels (MA 32), a product of rational quadratic kernels (RQ), and a product of spectral mixture kernels (SM). We use as boundary times for string GP kernels $\{0, 0.5, 1\}$. Figures \ref{fig:pred_f2} and \ref{fig:pred_f3} illustrate the ground truth (top left corner), the training data (top right corner), along with posterior means under the aforementioned kernels on $f_2$ and $f_3$ respectively. It can be seen  that a product of string periodic kernels considerably outperforms alternatives in both cases. Competing kernels tend to fuse local  patterns in the interpolation, whereas \textit{string GP kernels} recover the underlying functions almost perfectly. Interestingly, the product of \textit{string GP kernels} almost perfectly recovers $f_3$, despite it not being a separable function.

\section{Proofs}
\subsection{Proof that the fully nonparametric model of \cite{Schmidt03} yields a stationary prior}
We want to prove that if $\textbf{d}$ is a Gaussian process with mean the identity function and covariance function invariant by translation, and if conditional on $\textbf{d}$ $(z_x)$ is a GP  with mean function that does not depend on $\textbf{d}$ and conditional covariance function $\textbf{k}(x, x^\prime) = f \big(\| \textbf{d}(x) - \textbf{d}(x^\prime) \|)$ where $f$ is positive definite, then $(z_x)$ has (unconditional) covariance function invariant by translation.

Firstly, we note that $\textbf{d}$ can be rewritten as $\textbf{d}(x) = x + d_c(x)$ where $d_c$ is a stationary GP with mean $0$. Hence, the conditional covariance can be rewritten as
\[\text{cov}(z_x, z_{x+h} ~\vert ~\textbf{d}) := \textbf{k}(x, x+h) =   f \big(\| -h + d_c(x) - d_c(x+h)\|).\]
By law of total covariance, 
\[\text{cov}(z_x, z_{x+h}) = \text{E}\big(f \big(\| -h + d_c(x) - d_c(x+h)\|)\big) + \text{cov}\big(\textbf{E}(z_x~\vert ~\textbf{d}), \textbf{E}(z_{x+h}~\vert ~\textbf{d})\big).\]

The first expectation is with respect to the law of $d_c(x) - d_c(x+h)$, which is the same as the law of $d_c(0) - d_c(h)$ by stationary of $d_c$. Hence, $\text{E}\big(f \big(\| -h + d_c(x) - d_c(x+h)\|)\big)$ does not depend on $x$. Moreover, as the conditional mean $\textbf{E}(z_x~\vert ~\textbf{d})$ does not depend on $\textbf{d}$, $\text{cov}\big(\textbf{E}(z_x~\vert ~\textbf{d}), \textbf{E}(z_{x+h}~\vert ~\textbf{d})\big)=0$, which concludes the proof.

\subsection{Proof that the fully nonparametric GPPM model of \cite{ggpm} yields a stationary prior}
We prove that if $f$ and $g$ are two independent mean zero stationary GPs, then $y(x)=f(x)\exp(g(x))$ is also stationary, and we derive its covariance function.

Firstly, we recall that covariance stationarity of Gaussian processes implies strong stationarity. Moreover, strong stationarity of $g$ implies strong stationarity of $\exp\big(g\big)$. Finally, as the product of strongly stationary processes is strongly stationary, $y$ is also strongly stationary, which implies it is also covariance stationary.

As for the covariance function of $y$, we note that $y$ is also mean zero as \[\text{E}\big(y(x)\big):= \text{E}\big(f(x)\exp(g(x))\big) = \text{E}\big(f(x)\big)\text{E}\big(\exp(g(x))\big)=0,\]
where the second equality results from the independence between $f$ and $g$.

Hence, \begin{align}
\text{cov}\big(y(u), y(v)\big) &= \text{E}\big(y(u)y(v)\big) \nonumber \\
&=\text{E}\big(f(u)f(v)\big)\text{E}\big(\exp(g(u)+g(v))\big) \nonumber \\
&= k_f(u-v)\exp\big(k_g(0)+k_g(u-v)\big),
\end{align}
where $k_f(u-v), ~ k_g(u-v)$ are the covariance functions of $f$ and $g$. The second equality results from the independence between $f$ and $g$ and the last equality follows from the moment generating function of a Gaussian.

\subsection{Proof of the existence of derivative Gaussian processes}
See Appendix B of our string Gaussian processes paper.

\subsection{Proof of the existence of string Gaussian processes}
See Appendix C of our string Gaussian processes paper.

\subsection{Proof of the lemma of the paper}
See Appendix G of our string Gaussian processes paper.

\section{Joint inference of a latent function and its gradient under a GP prior}
Let us consider the following GP regression problem:
\[\forall ~ x \in \mathbb{R}^d, ~ y(x) = f(x) + \epsilon(x), ~~ f \sim \mathcal{GP}(0, k(.,.)), ~~ \epsilon(x) \overset{\text{i.i.d}}{\sim} \mathcal{N}(0, \sigma^2).\]
It can be shown in a similar fashion to the proof of the existence of derivative Gaussian processes that if $k$ is $\mathcal{C}^2$, then $f$ is $\mathcal{C}^1$ in the $L^2$ sense and $(f, \nabla f)$ is a $\mathbb{R}^{d+1}$ valued Gaussian process. Moreover, we have:
\begin{equation}
\text{cov}\big(y(u), \frac{\partial f}{\partial x_j}(v)\big)=\text{cov}\big(f(u), \frac{\partial f}{\partial x_j}(v)\big)=\frac{\partial k}{\partial v_j}(u,v),
\end{equation}
\begin{equation}
\text{cov}\big(\frac{\partial f}{\partial x_i}(u), \frac{\partial f}{\partial x_j}(v)\big)=\frac{\partial^2 k}{\partial u_i\partial v_j}(u,v),
\end{equation}
where $\frac{\partial f}{\partial x_j}(v)$ denotes the partial derivative of the latent function with respect to the $j$-th input dimension at $v$, $\frac{\partial k}{\partial v_j}(u,v)$ denotes the partial derivative of the covariance function with respect to the $j$-th coordinate of the second input evaluated at $(u, v)$, and $\frac{\partial^2 k}{\partial u_i\partial v_j}(u,v)$ denotes the second order partial derivative of the covariance function with respect to the $i$-th coordinate of the first input and the $j$-th coordinate of the second input evaluated at $(u, v)$. 

When $k$ is separable (as in the global air temperature anomalies) and $k(u, v) = \prod_{j=1}^{d} k_j(u_j, v_j)$, it follows that \[\frac{\partial k}{\partial v_j}(u,v)= \frac{\partial k_j}{\partial v_j}(u_j,v_j)\prod_{l \neq j} k_l(u_l, v_l),\] 
\[\forall i \neq j, ~\frac{\partial^2 k}{\partial u_i\partial v_j}(u,v)=\frac{\partial k_i}{\partial u_i}(u_i,v_i)\frac{\partial k_j}{\partial v_j}(u_j,v_j)\prod_{l \notin \{i, j\}} k_l(u_l, v_l)\] 
and 
\[\frac{\partial^2 k}{\partial u_i\partial v_i}(u,v)=\frac{\partial^2 k_i}{\partial u_i \partial v_i}(u_i,v_i)\prod_{l \neq i} k_l(u_l, v_l).\]
We recall that when all univariate kernels in the product are \textit{string GP kernels}, $\frac{\partial k_j}{\partial v_j}(u_j,v_j)$, $\frac{\partial k_i}{\partial u_i}(u_i,v_i)$ and $\frac{\partial^2 k_i}{\partial u_i \partial v_i}(u_i,v_i)$ are elements of the matrix $\bar{\textbf{K}}_{u;v}$ derived in the paper.

Overall, the posterior distribution of the gradients at test points given noisy measurments of the latent function at training points is Gaussian and its mean and covariance matrix are obtained using standard Gaussian identities (\cite{rasswill}) and the joint covariance structure we just derived.
\end{appendices}
\end{document}